\documentclass[a4paper,fleqn]{cas-dc}

\usepackage[numbers]{natbib}
\usepackage{amsmath,amssymb,amsfonts}
\usepackage{algorithmic}
\usepackage{graphicx}
\usepackage{textcomp}
\usepackage{multirow}
\usepackage{subfigure}
\usepackage{gensymb} 
\usepackage{url}
\usepackage{amssymb}
\usepackage{comment} 
\usepackage{caption}
\usepackage{stackengine}
\usepackage{pbox}
\usepackage{cas-common}

\def\tsc#1{\csdef{#1}{\textsc{\lowercase{#1}}\xspace}}
\tsc{WGM}
\tsc{QE}
\newcommand\barbelow[1]{\stackunder[1.2pt]{$#1$}{\rule{.8ex}{.075ex}}}
\def\ubx{{\barbelow{\ensuremath{\mathbf x}} }}
\def\ubA{{\barbelow{\ensuremath{\mathbf A}} }}

\def\ccalD{{\mathcal D }}
\def\ccalS{{\mathcal S }}
\def\ccalX{{\mathcal X }}

\def\reals{{\mathbb R}}

\def\barbX{\bar{\mathbf X}}

\def\bbalpha{\boldsymbol{\alpha}}

\begin{document}
\let\WriteBookmarks\relax
\def\floatpagepagefraction{1}
\def\textpagefraction{.001}

\shorttitle{Multimodal Interpretable Data-Driven Models for Early Prediction of AMR using MTS}    

% \shortauthors{<short author list for running head>}  

\title [mode = title]{Multimodal Interpretable Data-Driven Models for Early Prediction of Antimicrobial Multidrug Resistance Using Multivariate Time-Series}  

% \tnotemark[<tnote number>] 



\author[1]{Sergio Mart{\'i}nez-Agüero}

% % Email id of the first author
\ead{sergio.martinez@urjc.es}

% % URL of the first author
% \ead[url]{<URL>}

% \credit{<Credit authorship details>}

\author[1]{Antonio G. Marques}
\ead{antonio.garcia.marques@urjc.es}
\author[1]{Inmaculada Mora-Jim{\'e}nez}
\ead{inmaculada.mora@urjc.es}
\author[2]{Joaqu{\'i}n Alv{\'a}rez-Rodr{\'i}guez}
\ead{joaquin.alvarez@salud.madrid.org}
\author[1]{Cristina Soguero-Ruiz}
\ead{cristina.soguero@urjc.es}

% \address[1]{Spain (e-mails: \{sergio.martinez, cristina.soguero, inmaculada.mora, antonio.garcia.marques\}@urjc.es)}

% \address[2]{, Fuenlabrada 28942, Spain (e-mail joaquin.alvarez@salud.madrid.org )}

% % Address/affiliation
\affiliation[1]{organization={Department of Signal Theory and Communications, Telematics and Computing Systems, Rey Juan Carlos University},
            addressline={}, 
            city={Fuenlabrada},
%          citysep={}, % Uncomment if no comma needed between city and postcode
            postcode={28942}, 
            state={Madrid},
            country={Spain}}
\affiliation[2]{organization={Intensive Care Department, University Hospital of Fuenlabrada},
            addressline={}, 
            city={Fuenlabrada},
%          citysep={}, % Uncomment if no comma needed between city and postcode
            postcode={28942}, 
            state={Madrid},
            country={Spain}}


\begin{abstract}
Electronic health records (EHR) is an inherently multimodal register of the patient's health status characterized by static data and multivariate time series (MTS). While MTS are a valuable tool for clinical prediction, their fusion with other data modalities can possibly result in more thorough insights and more accurate results.  Deep neural networks (DNNs) have emerged as fundamental tools for identifying and defining underlying patterns in the healthcare domain.  However, fundamental improvements in interpretability are needed for DNN models to be widely used in the clinical setting.  In this study, we present an approach built on a collection of interpretable multimodal data-driven models that may anticipate and understand the emergence of antimicrobial multidrug resistance (AMR) germs in the intensive care unit (ICU) of the University Hospital of Fuenlabrada (Madrid, Spain). The profile and initial health status of the patient are modeled using static variables, while the evolution of the patient’s health status during the ICU stay is modeled using several MTS, including mechanical ventilation and antibiotics intake. The multimodal DNNs models proposed in this paper include interpretable principles in addition to being effective at predicting AMR and providing an explainable prediction support system for AMR in the ICU. Furthermore, our proposed methodology based on multimodal models and interpretability schemes can be leveraged in additional clinical problems dealing with EHR data, broadening the impact and applicability of our results.

\end{abstract}

% Keywords

\maketitle

% Main text

\section{Introduction}
\label{sec:introduction}

% Multimodal Data
Data-driven machine learning (ML)
% \textcolor{red}{methods? para hacerlo más genérico?} 
% classifiers 
methods have emerged as crucial tools in healthcare applications. The most common way to collect data in the clinical setting is through Electronic Health Records (EHR), a record of patients' health status and evolution. EHR data are naturally multimodal, with each patient having diverse and complementary information represented by variables of different nature that capture his/her health status. The different types (modalities) of information include, among others, binary and continuous static demographic data, categorical health-status data 
% (typically coded as small binary vectors)
, or more complex time-varying measurements that need to be modeled as multivariate time series (MTS). Albeit more 
% difficult 
challenging to process, the last decade has witnessed a growing interest in analyzing clinical data as time-series sequences, allowing clinical experts to assess better the patient health evolution~\cite{funkner2017data, ghassemi2015multivariate, tatonetti2012data}.

%\textcolor{blue}{CSR: esta idea creo que está en la línea que empieza por furthrermore, revisar. SMA: He fusionado ambas frases.}
While MTS are indeed a valuable tool in clinical prediction, its fusion (i.e., joint consideration) with other data modalities can provide a holistic picture of patient status, potentially leading to more comprehensive insights, more precise results, more reliable behaviors, stronger acceptance from the medical community~\cite{MORADI2015398, qiao2019mnn, nagpal2017deep}. Furthermore, joint consideration of multiple data modalities is effective in reducing noise by obtaining complementary information from different data sources~\cite{zhang2020advances}. For all these reasons, in recent years, several works in the clinical context have looked at the application of multimodal data science and ML architectures that combine inputs of different types (including static features and MTS) to generate enhanced and more comprehensive clinical predictions. Within this line of works, Cheng et al. applied a set of deep fusion neural networks (NNs) to predict gastrointestinal bleeding hospitalizations based on different multimodal data recorded in the EHR~\cite{hung2019predicting};  Shuai et al. used a fusion classifier with attention mechanisms to predict the disease risk using text notes and MTS~\cite{niu2021label}; and Li et al. developed a multimodal model to integrate information on demographics, medical notes, and clinical MTS~\cite{li2021integrating}.

Given the complexity and irregular patterns present in real clinical datasets, deep NNs (DNNs) have emerged as a valuable resource to characterize and find the underlying relationships in MTS~\cite{lecun2015deep, shickel2017deep}. 
One of the most widely-used deep learning approaches for dealing with time-series sequences is the Gated Recurrent Unit (GRU)~\cite{shickel2017deep, cho2014learning}. The GRU is a modification of standard Recurrent NNs (RNNs) widely employed to deal with MTS due to their capability of using time-varying observations and learning long-term temporal dependencies~\cite{bahdanau2015mtranslation}.

Although the effectiveness of deep learning models has been proven in the literature~\cite{shickel2017deep, piccialli2021survey}, the performance improvement comes with a cost: models are so complex that underlying mechanisms are too difficult to capture, and only indirect analysis can be applied to gain insights into the role of the different input features~\cite{goh2017smiles2vec}.  The lack of interpretability in deep learning models is currently the main barrier to applying such powerful models in the medical context to support clinical decision-making based on understandable relationships~\cite{london2019artificial}.

Wide adoption of deep learning models in the clinical context requires fundamental advances in  ML interpretability~\cite{sendak2020human}. Consequently, in recent years, a multitude of interpretable models have emerged in the healthcare domain based on different methods, including: (i) Feature importance methods~\cite{shrikumar2017learning, molnar2020interpretable}; (ii) feature interaction attribution~\cite{janizek2021explaining, crabbe2021explaining}; (iii) neuron layer attribution~\cite{dhamdhere2018important, shrikumar2018computationally}; and (iv) explanation with high-level concepts~\cite{ghorbani2019towards, martinez2022interpretable}, to name a few.

In this work, we propose a methodology based on a set of \emph{interpretable multimodal data-driven models} capable of predicting and grasping knowledge about the emergence of Antimicrobial Multidrug Resistance (AMR) in the Intensive Care Unit (ICU). AMR can be characterized as the capacity of microorganisms to withstand the impacts of an assortment of harmful chemical agents intended to damage them~\cite{michael2014antimicrobial}. 
The adaptation of the bacteria to different antimicrobials (to which they were previously sensitive) hinders the treatment of the infection, worsening the patients' conditions and reducing the range of secondary antimicrobials available~\cite{michael2014antimicrobial, magiorakos2012multidrug}. As a result, situations such as cuts, care of premature babies, chemotherapy against cancer, or infections can cause debilitating or even lethal outcomes~\cite{michael2014antimicrobial, infectious2011combating}.

In a nutshell, this work proposes the joint use of irregular MTS, demographic features, and interpretable mechanisms to gain insights and predict the ICU AMR onset.
Previous works by the group have used ML and data-based tools for predicting AMR onset~\cite{martinez2022interpretable, pascual2021predicting, escudero2021use, arnanz2020feature, rey2020feature, escudero2020temporal} considering each time instant separately without fully exploiting the temporal variations and similarities among patients. 
In contrast, this paper: i) puts forth irregular time series models able to capture inter and intra dependencies of MTS and ii) combines that information with the one contained in non-MTS demographic features.
The methodology and data-science pipeline proposed here can be used by clinicians as a data-based tool to help in the discovery and understanding of the development and spreading of AMR germs in the ICU.  Our main contributions are the following:

\begin{itemize}
    \item Analyzing and modeling MTS and static features related to AMR in the challenging scenario of an ICU. The dataset contains data representing the health status of 3,470 patients. To obtain as much information as possible from the data, a cleaning and modeling process has been performed. Also, we developed methods to solve problems specific to AMR classification, such as population unbalance, MTS irregularity, or high dimensionality of the data. 
     
    \item Developing multimodal architectures to characterize the patient's initial status and evolution. To characterize the emergence of AMR germs, we have used the static features to model the initial health status of the patient, then the evolution of the patient's health status is modeled by MTS. The best results have been obtained with the ``First Hidden State Initializer'' architecture, a sample-dependent variable selection model that creates an encoding vector to provide extra context to the MTS. 
    
    \item Regarding knowledge extraction, we have applied two complementary approaches: Feature Selection (FS) and interpretable mechanisms. We first studied the effect of classical FS methods. Then, we used a permutation multimodal FS approach. We have evaluated both FS procedures in terms of performance and interpretability, thus finding relevant features. Finally, we applied different interpretable mechanisms to learn hidden patterns
    present within the dataset.
    
\end{itemize}

The remainder of the paper is organized as follows. Sec.~\ref{sec:Methods} presents the notation and methods used in this work. Sec.~\ref{sec:Database} describes the dataset and the related pre-processing tasks. Experiments and results are provided in Sec.~\ref{sec:Experiments}. The main conclusions and %associated
the discussion are drawn in Sec.~\ref{sec:Conclusion}.

\section{Methods}
\label{sec:Methods}

\begin{figure}[ht!]
\centering
	\includegraphics[width=\columnwidth]{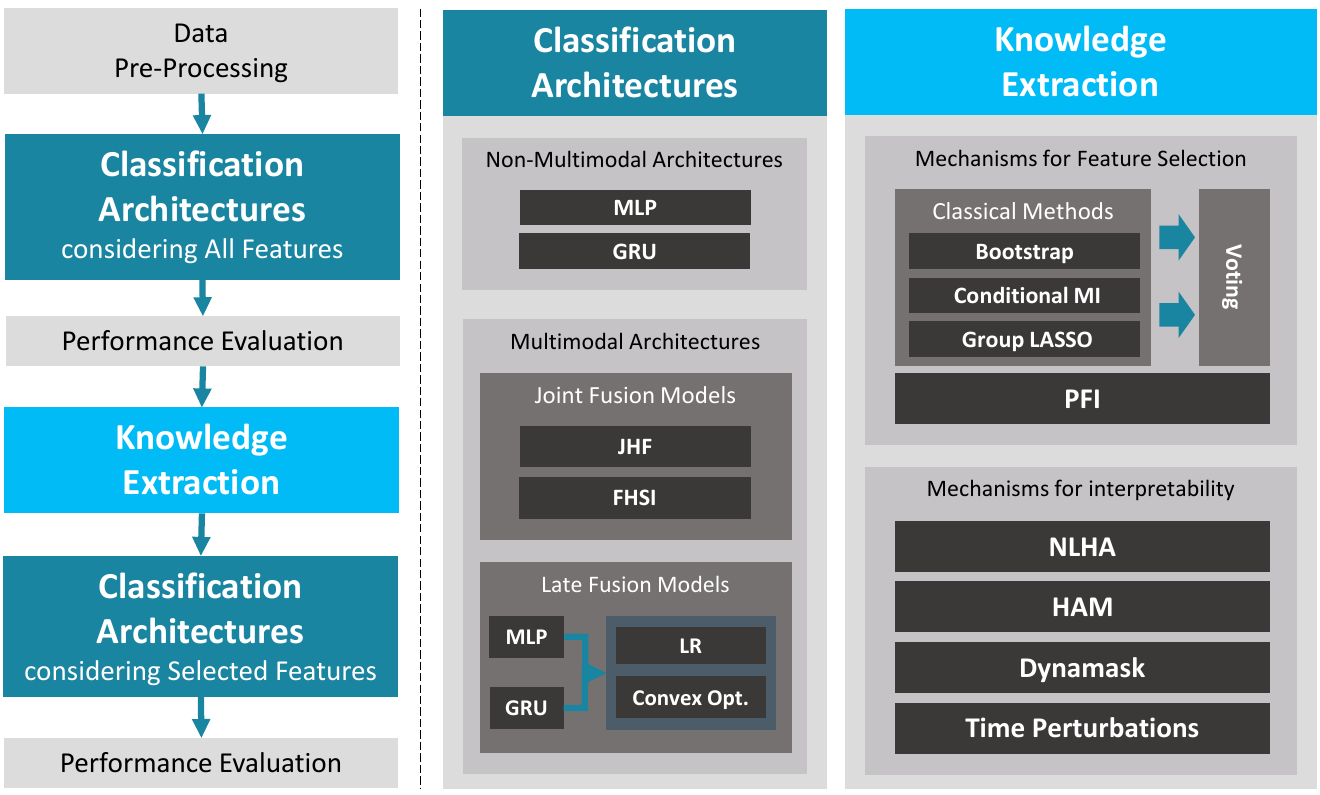}
	\caption{
	Graphical illustration of the workflow implemented. 
	As illustrated in the left column, we begin by running a pre-processing stage to promote consistent and reliable results. 
	Then, non-multimodal and multimodal models using all available features are trained (see Sec.~\ref{sec:NN} and central column). 
	We perform a knowledge extraction step for studying the most important features and time-slots, using two different FS schemes (see Sec.~\ref{subsec:FeatureSelection} and right column) and different interpretable models (see Sec.~\ref{subsec:Interpretable_methods} and right column). 
	Once the most important variables are selected, we train models using the knowledge acquired using the FS and interpretable schemes. Finally, the models' performance and interpretability are evaluated using several figures of merit (see Sec.~\ref{sec:Experiments}).
    }
	\label{fig:workflowFramework}
\end{figure}

The experimental pipeline followed in this work is sketched in Fig.~\ref{fig:workflowFramework} and discussed in the following subsections.
Data pre-processing and mathematical notation are introduced in Sec.~\ref{sec:prep}. 
Sec.~\ref{sec:NN} describes the DNN architectures designed to perform the prediction of the AMR.
Multimodal and fusion strategies are described in Sec.~\ref{Sec:MethodsMultimodal}. Finally, the methods used for knowledge extraction are presented in Sec.~\ref{subsec:FeatureSelection}, and Sec.~\ref{subsec:Interpretable_methods}.\footnote{The ML and data processing architectures developed in this paper have been programmed in Python. The associated code is publicly available at \url{https://github.com/smaaguero/MIDDM}}.

\subsection{Preliminaries}
\label{sec:prep}

The first step is to gather and process the clinical information of the different patients. For this section, it suffices to say that the collection of the data has been described before~\cite{martinez2022interpretable}. 
% and that, 
Moreover, as preliminary pre-processing tasks, we have implemented normalization, database homogenization, and outliers treatment, which are all critical when dealing with real clinical databases~\cite{catley2008multi, khazaei2014toward}. A more detailed description of the pre-processing stage is provided in Sec.~\ref{sec:Database}, once the notation, problem statement, and ML methods have been introduced.

The second goal of this subsection is to introduce the mathematical notation used throughout the manuscript. We consider $I$ patients, indexed by $i=1,2,...,I$. 
Since we are dealing with multimodal data, the data associated with the patient (say the $i$-th one) is collected into two different mathematical variables: the $\mathbf{X}_i$ matrix, which represents the MTS data, and the $\mathbf{z}_i$ vector, which represents the static data.
\begin{itemize}
\item The input matrix $\mathbf{X}_i$ of each patient is formed as a collection of $D$ time series, all of them with length (duration) $T_i$. We emphasize that the value of $T_i$ depends on the time the patient $i$ stayed in the ICU. Therefore, data associated with the $i$-th patient can be arranged in the matrix  $\mathbf{X}_i\in\mathbb{R}^{D \times {T_i}}$. 
To simplify some of the mathematical expressions in later sections, we denote the entries of matrix $\mathbf{X}_i$ as $x_i^{(t,d)}$, with the latter representing the value of the $d$-th time series variable in the $t$-th time-slot for the $i$-th patient. We then define the vectors $\bar{\mathbf{x}}_i^t$ and $\ubx_i^d$, where vector $\bar{\mathbf{x}}_i^t$  contains the $D$ features associated with the $i$-th patient during the $t$-th time-slot, so that $\bar{\mathbf{x}}_i^t =[x_i^{(t,1)}, x_i^{(t,2)}, \cdots, x_i^{(t,D)}]^\top$, and, analogously,  vector $\ubx_i^d=[x_i^{(1,d)}, x_i^{(2,d)}, \cdots, x_i^{(T_i,d)}]^\top$ collects the $T_{i}$ values of the $d$-th feature of patient $i$.
\item  The input vector $\mathbf{z}_i$ corresponding to the static feature is formed by a set of $G$ values, each value corresponding to a feature. Hence, $z_i^{g}$ represents the value of the $g$-th variable for the $i$-th patient.
\end{itemize}
 
Regarding the output of our ML architecture, we cast the clinical problem of AMR identification as a binary classification task. Therefore the label `1' is used to identify patients for whom an AMR germ has been detected, and the label `0' is used to identify the non-AMR ones. We denote the label associated with the $i$-th patient $y_i\in\{0,1\}$, and the output generated (predicted) by the ML model at hand as $\hat{y}_i$ (depending on the model, $\hat{y}_i$ will be either a binary or a real value between 0 and 1).

\subsection{Non-Multimodal NN Architectures}
\label{sec:NN}

NNs are ML approaches widely used to handle (clinical) data due to their ability to unveil complex non-linear dependencies~\cite{Duda_01}. These methods can be applied for regression and classification tasks and deal with data of different nature.

In this paper, we focus on a binary classification task, firstly using non-multimodal (NM) architectures: a simple multilayer perceptron (MLP) for dealing with static data~\cite{su1993integration} and a more sophisticated RNN when considering MTS ~\cite{Graves12}. Furthermore, more complex multimodal architectures with the capacity to analyze both static data and MTS are built.

\subsubsection{Multilayer Perceptron for Processing Static Data}
An MLP is an NN designed as the concatenation (successive application) of a collection of layers formed by a set of neurons~\cite{sordo2002introduction}.  
 
Each layer is composed of several (parallel) neurons, and each neuron implements a \emph{non-linear} (activation) function that generates a unidimensional output using as input a unidimensional value found as the \emph{linear} combination of multiple inputs using a set of weights. The weights performing the linear combinations are adjusted during the learning process, which is performed by optimizing a cost function using stochastic gradient-based approaches~\cite{lecun2015deep}. The ease of use and the capability of being a universal classifier have converted the MLP into one of the most widely used architectures in problem-solving \cite{ripley2007pattern}.

\subsubsection{Gated Recurrent Unit for Processing Temporal Data}

RNNs are a type of NNs that, due to their internal representation of a state-space model, are specialized in dealing with temporal data, including MTS. 
While traditional DNNs deal with each of the time instants of the MTS separately, the RNN accounts for time-dependencies by employing an internal state that provides an `artificial memory' of the previous inputs. However, RNNs cannot reach their full potential in applications where long MTS are involved (as is the case in this work) since the application of gradient steps that either decay or blow exponentially (see, e.g., ~\cite{Graves12}, for more details on the so-called vanishing gradient problem).

GRUs are a modification of the standard RNNs featuring a gating mechanism aimed at bypassing the vanishing gradient's problems~\cite{Cho_2014}. 
A ``gate'' is a structure whose purpose is to regulate the flow of information going along the network, deciding which information contained in the MTS is important to keep or throw away. A gating mechanism can perform different tasks, such as amplifying a vanishing gradient or guaranteeing that the error goes through. The GRU has two mechanisms to regulate the information: i) the reset gate eliminates the redundant information contained in the previous hidden state, keeping only the relevant information of previous time-slots; ii) the update gate obtains the relevant information contained in the current time-slot. Then, both the information obtained from the previous hidden state (output of the reset gate) and the information obtained from the current time-slot (output of the update gate) are combined~\cite{dey2017gate}. GRU networks require fewer parameters than other RNNs, and this is a desirable property in clinical applications, where the number of samples is typically limited. 

\subsection{Multimodal DNN Architectures}\label{Sec:MethodsMultimodal}

``Data fusion'' (aka ``Multimodality'') refers to combining data from multiple modalities to extract complementary and more accurate and comprehensive knowledge~\cite{gao2020survey, meng2020survey}. Depending on the combination approach, we can differentiate three different data fusion families: early fusion, joint fusion, and late fusion~\cite{huang2020fusion}. 
\emph{Early fusion} models combine the input from different modalities before feeding the model. Input modalities can be combined in different ways, including concatenation, pooling, or applying a gated unit~\cite{zhang2020advances}.
\emph{Joint fusion} is the process of combining different feature mappings generated by the intermediate layers of the architecture. In joint fusion architectures, the loss is propagated back to the feature extracting NNs during training, giving rise to updated feature mappings that enhance the original mappings created by the early fusion layers~\cite{huang2020fusion}.
Differently, \emph{late fusion} architectures combine \emph{predictions} from multiple models to generate a single prediction. For the case of late fusion classification architectures dealing with dynamic and static data, the architecture can be divided into three blocks: one devoted to generating a posteriori probability from static data, one devoted to generating a posteriori probability from MTS, and one that combines both probabilities to generating the label estimated by the late fusion model. In the healthcare domain, integrating static information such as age or comorbidities with MTS is very important from a clinical point of view to support clinical decision-making. As a result, several data-fusion architectures have been recently proposed in the clinical context~\cite{hung2019predicting, niu2021label, li2021integrating}.

Building upon the structure of a GRU, we have carefully designed three novel multimodal architectures to deal with MTS and static data. 
For the setup at hand, adopting early fusion architectures would require treating the static features as time series repeating the static variables over time.  
Therefore, we have focused on \emph{joint fusion} (``Joint Heterogeneous Fusioner'' and ``First Hidden State Initializer'') and \emph{late fusion} architectures (``Late Fusion Convex Optimization'' and ``Late Fusion Logistic Regression''). 
All the architectures listed above will be explained later in this work.

\subsubsection{Joint Heterogeneous Fusioner}
As introduced above, there are many types of data fusion architectures. Often, it is reasonable to assume that the different data fusion modalities do not independently affect the target but rather that informative cross-modality interactions exist. In joint fusion, such relationships are modeled by learning interactions of features from the intermediate representations. These interactions can be learned by first concatenating the marginal representations and feeding this vector into fully connected layers before a task-specific output layer~\cite{Stahlschmidt2022Bioinformatics}. 

In this work, we have designed the Joint Heterogeneous Fusioner (JHF), an architecture that creates two different representations using the static features and the MTS. In our design, the intermediate layers take advantage of the ``prior knowledge'' we have about the structure of the modality variables. We have used a GRU to identify and model the interactions between the MTS, summarizing the information into a vector. Similarly, we have employed the broadly used entity embeddings~\cite{gugulothu2017predicting} for categorical static variables as feature representations and linear transformations for binary and numeric static variables. A wide range of methods exist for unifying marginal representations; in this work, we have chosen concatenation due to its wide adoption and ease of interpretation. Finally, we apply a linear transformation layer followed by a sigmoid activation function over the concatenated representations.

\subsubsection{First Hidden State Initializer}
Temporal fusion transformers (TFT), a complex architecture with multiple innovations, have been shown to yield significant performance improvements over state-of-the-art benchmarks in time series forecasting using static data and MTS~\cite{LIM20211748}. Motivated by this, we leverage the original TFT, modifying it to account for the structure of our setup and giving rise to a joint fusion multimodal architecture referred to First Hidden State Initializer (FHSI).

\begin{figure}[ht!]
\centering
	\includegraphics[width=\columnwidth]{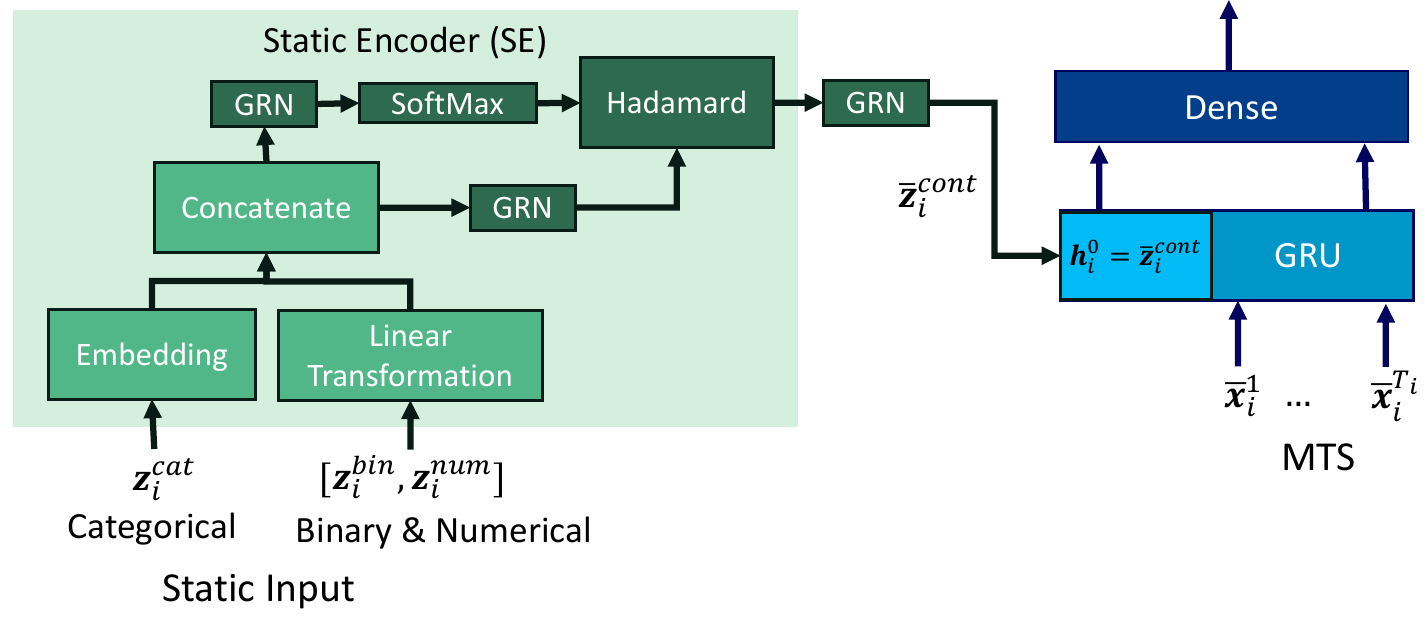}
	\caption{High-level architecture of the FHSI. FHSI deals with static and time-varying inputs. The different blocks of the architecture are represented using different colors. The SE is represented in different green colors; the light green color blocks represent a first embedding mapping network, and the dark green color blocks represent the Variable Selection Network. 
	The GRU block is represented in a light blue box, and the last non-linear dense layer is represented in a dark blue box.}
	\label{fig:FHSI_architecture}
\end{figure}

In the clinical context, knowing the initial status of a patient is crucial to understanding the patient's evolution. 
% This initial state has a significant impact on the medications and procedures that the patient undergoes during his/her stay. 
This initial state significantly impacts the medications and procedures the patient undergoes during his/her stay. 
Following this idea, the FHSI architecture uses static features to create a context vector that enriches the first hidden state of a GRU. To generate such a context vector, the FHSI scheme uses an internal module named Static Encoder (SE). Figure~\ref{fig:FHSI_architecture} shows the high-level architecture of the FHSI, with individual components detailed in different colors. 
\begin{itemize}
\item To build the context vector (denoted as $\bar{\mathbf{z}}_i^{cont}$), the SE first implements a mapping (embedding) for the different static features. Since we are dealing with categorical, binary, and numerical features, we have employed different strategies to build this first vector representation.

For the categorical variables $\mathbf{z}_i^{cat}$, we have employed the broadly used entity embeddings~\cite{gugulothu2017predicting} as feature representations and linear transformations for binary and numeric variables $\mathbf{z}_i^{bin}$ and $\mathbf{z}_i^{num}$. Note that this first mapping is represented in a light green color in Figure~\ref{fig:FHSI_architecture}.
\item 
The next block within the SE implements a variable selection mechanism (represented in dark green color in Figure~\ref{fig:FHSI_architecture}). 
The variable selection mechanism creates a vector for each patient, weighting the original input using a Hadamard product~\cite{horn1990hadamard}. With the motivation of endowing the model with the flexibility to apply non-linear processing only where needed, we propose using the well-known Gated Residual Network (GRN) as a building block in the variable selection network~\cite{LIM20211748, tan2018gated}. The output of the SE is denoted as $\bar{\mathbf{z}}_i^{cont}$.
\item Finally, we use the generated context vector $\bar{\mathbf{z}}_i^{cont}$ as the initial state of the GRU (represented in light blue in in Figure~\ref{fig:FHSI_architecture}), which is in charge of dealing with the MTS $\mathbf{X}_i=[\bar{\mathbf{x}}_i^1, \bar{\mathbf{x}}_i^2,\ldots,\bar{\mathbf{x}}_i^{T_i}]$. Specifically, the GRU has $T_i+1$ internal (hidden) states $\mathbf{h}_i^t$ each of them associated with the corresponding $\bar{\mathbf{x}}_i^t$ plus one additional initialization state. As explained before, our proposed architecture sets the initial hidden state to the output of the SE as $\mathbf{h}_i^0=\bar{\mathbf{z}}_i^{cont}$.
\end{itemize}

\subsubsection{Late Fusion Models}
\label{subsec:late_fusion_method}
In many applications, ensemble learning approaches that combine (aggregate) the outputs of multiple simple models lead to a better performance than that achieved by the individual models~\cite{sanchez2008adaptive}.
The most widely-accepted procedure is to train the basic models, add a module to aggregate the outputs and retrain to estimate the values of the aggregation parameters. The aggregator can be based on different approaches,  such as parameter optimization, weighting coefficients, or error-processing techniques~\cite{ren2014optimal}.

In the context of late fusion for multimodal data, a sensitive design approach is to define one model for static data, one model for MTS data, and add an aggregator to combine the two separate outputs into a single one. This is indeed the approach implemented in this paper, where two late fusion architectures are considered. In both cases, we use an MLP to deal with static data and a GRU to deal with MTS, with each of them returning a value for the a posteriori probability of developing an AMR infection. The difference is in the aggregation module, where two simple alternatives are considered. One of them implements a linear combination, and the other one a non-linear one, as described next.  

\begin{itemize}
    \item The first late fusion approach is referred to as Late Fusion Convex Optimization (LFCO). This model performs a linear combination of the individual MLP and GRU models using two different weights, $w_{MLP}\in[0,1]$ and $w_{GRU} \in[0,1]$. Since the combination is constrained to be convex, the constraint $w_{GRU} + w_{MLP} = 1$ must be enforced. The optimal values for $w_{GRU}$ and $w_{MLP}$ are found by implementing a simple unidimensional exhaustive search aimed at maximizing the classification performance over the validation test. On top of its simplicity, one additional advantage of the LFCO approach is its ease of interpretation, with the value of the ratio $w_{GRU}/w_{MLP}$ representing the relative importance of the dynamic variables relative to the static variables for the prediction of the label.
    
    \item The second approach, referred to as Late Fusion Logistic Regression (LFLR), employs an LR to aggregate the MLP and GRU outputs. 
    Note that the LR is only concerned with merging the outputs of the MLP and GRU since it is working as aggregator. Therefore, during the LR training process, neither the MLP nor the GRU are retrained.
    The LR is a widely used parametric approach that estimates the final prediction value by applying a non-linear logistic function to a linear combination of the input features (the MLP and GRU outputs in our case)~\cite{tolles2016logistic}. Note that the LR is only responsible for the fusion of the MLP and GRU outputs. Therefore, when learning (training) the LR, neither the MLP nor the GRU are retrained.    
\end{itemize}
 
\subsection{Mechanisms for FS}
\label{subsec:FeatureSelection}
Real-world clinical data oftentimes contain irrelevant, redundant, and noisy features. Following an FS approach is essential to enhance classification performance, avoid loss of information, and increase generalization~\cite{tang2014feature, munoz2020informative}.
In applications where the number of patients is limited, the use of FS techniques is even more crucial since the reduction of the dimensions of the inputs entails that the architectures need to learn a smaller number of parameters. In addition, FS provides a disciplined data-driven strategy for identifying the most relevant features for the task at hand, providing insights on the problem, and enhancing the interpretability of models.

For completeness, the following subsections discuss the four FS techniques implemented in this paper. The first three correspond to classical FS schemes in statistics: Confidence Intervals with Bootstrap (CIB)~\cite{efron1982jackknife, efron1994introduction}, Conditional Mutual Information (CMI)~\cite{li1990mutual} and Group Least Absolute Shrinkage and Selection Operator (GLASSO)~\cite{chesneau2008some}, with one of the objectives in the exposition being the description of how those techniques can deal with MTS. We finally introduce a method for FS based on already trained models called \textit{Permutation Feature Importance} (PFI)~\cite{breiman2001random, breiman2001statistical}. Readers familiar with FS can skip the remainder of the  subsection and move directly to Sec.~\ref{subsec:Interpretable_methods}.

\vspace{1mm}
\noindent\textbf{Confidence Intervals with Bootstrap:}~
Bootstrap resampling is a non-parametric strategy used to assess the distribution of a statistic (e.g., the median value) by taking random samples from a specific population~\cite{efron1982jackknife}. Bootstrapping does not make any assumption on the actual distribution function beyond the consideration that the observed and actual distributions are not dissimilar, which is suitable when the actual distribution is unknown~\cite{efron1994introduction}.

In our work, we use bootstrap resampling to evaluate whether the values of a variable in the AMR population are significantly different from the values of the same variable in the non-AMR population through a hypothesis test.  The associated feature is preserved if the variable is deemed sufficiently different.
To be more specific, we denote the population of AMR patients as $\ccalS_{AMR} $ and the set of non-AMR patients as $\ccalS_{non-AMR}$. The first step to perform the hypothesis test is to compute the difference between $\mu_{AMR}$ (the mean value of a feature in the population $\ccalS_{AMR}$) and $\mu_{non-AMR}$ (the mean of the same feature in the population $\ccalS_{non-AMR}$. The second step is to determine if the difference $\Delta =\mu_{AMR}-\mu_{non-AMR}$ is significant.
In order to implement a statistically robust procedure, we compute the resampling bootstrap approach rather than computing a simple and deterministic $\Delta$ using all patients in $\ccalS_{AMR}$ and $\ccalS_{non-AMR}$. Hence, we resample each population $R$ times, obtaining the sets $\{\ccalS^{(r)}_{AMR}\}_{r=1}^R$ for AMR patients and $\{\ccalS^{(r)}_{non-AMR}\}_{r=1}^R$ for non-AMR ones. 

The FS method based on CIB assumes that the features are unidimensional scalars. Therefore, CIB is directly applicable to the numerical static variables in $\mathbf{z}_i$. However, applying CIB is not straightforward for MTS. Given the patient-data matrices $\mathbf{X}_i$ and focusing on a particular time series (say the $d$-th one), we have to decide whether to keep or remove the $d$-th row of the data matrices for all the patients in the dataset. In other words, for each $d=1,...,D$, we need to determine if the $T_{i}$-dimensional vectors $\{\ubx_i^d\}_{i=1}^I$ are selected to be part of the inputs provided to our ML architectures. The handle this, for each feature (say the $d$-th one), we first run $T_{i}$ hypothesis tests to assess if each of the $t$-th entries of the vector $\ubx_i^d$ is individually relevant. Then, we implement a majority-rule scheme where the  $d$-th feature is selected if more than half of the individual time instants are considered relevant.

\vspace{1mm}
\noindent\textbf{Conditional Mutual Information:}~
The approach, in this case, is to implement an FS scheme so that the CMI between the selected features and the label $y$ is maximized. The concept of CMI is related to the Shannon entropy~\cite{li1990mutual}. 
To be mathematically precise, with $\ccalX$ denoting the set of values the (discrete) random variable $X$ can take, the entropy of $X$ is defined as $\mathbb{H}(X) = - \sum_{x \in \mathcal{X}} p(x) log(p(x))$, where $p(x)$ is $Pr\{X=x\}$. When two random variables ($X$ and $Y$) are present, two different generalizations of entropy can be defined. One is the  joint entropy, which is defined as $\mathbb{H}(X,Y) = - \sum_{x \in  \mathcal{X}} \sum_{y \in  \mathcal{Y}} p(x,y) log (p(x,y))$, with~$p(x,y)=Pr\{X=x, Y=y\}$. The second one, which is the most relevant one in the context of FS, is the conditional entropy, which is defined as
\begin{equation}
\mathbb{H}(X|Y)= - \sum_{x \in  \mathcal{X}} \sum_{y \in  \mathcal{Y}} p(x,y) log (p{(x|y})),
\end{equation}
with $p(y|x)=Pr\{Y=y|X=x\}=Pr\{X=x,Y=y\}/Pr\{X=x\}$. The MI between $X$ and $Y$ measures the shared information between both variables and is expressed as 
\begin{equation}
\mathbb{I}(X,Y)=\mathbb{H}(X)-\mathbb{H}(X|Y) = \mathbb{H}(Y) - \mathbb{H}(Y|X) = \mathbb{I}(Y,X).~~
\end{equation}
More specifically, the MI above quantifies the amount of information the variable $X$ has about the variable $Y$. 

With all this notation at hand, we are ready to define the CMI as the expected value of the MI of two random variables given a third random variable~\cite{gao2015efficient, fleuret2004fast}, so that 
\begin{equation}
\mathbb{I}(X,Y|Z)= \mathbb{H}(X,Z) - \mathbb{H}(Y|Z) - \mathbb{H}(X,Y,Z) - \mathbb{I}(Z).
\end{equation}
CMI is a widely-used metric for carrying out FS. The goal of CMI-based FS is to obtain the set $\ccalD' \subseteq \{1,2,...,D\}$ of $D'$ features that maximize the CMI between the reduced input $\mathbf{X}^{{\ccalD'}}$ and the associated label $y$. 
Solving that optimization exactly incurs exponential complexity, and, to bypass this, we implement an iterative unidimensional optimization of the CMI metric that, at each iteration, selects the most informative feature not yet present in $\ccalD'$. Furthermore, when estimating the value of $\mathbb{I}\big(y,\ubx^d~\big|\{\ubx^{d'}\}_{d'\in \ccalD'}\big)$ from the populations, we need to account for the fact that the variables $\ubx^d$ are multi-dimensional (so that $\ccalX$ is the Cartesian product of the value sets for each of the entries of $\ubx^{d'}$).

\vspace{1mm}
\noindent\textbf{Group LASSO:}~
The Least Absolute Shrinkage and Selection Operator (LASSO) is a well-known statistical method to regularize regression and classification problems that, as a byproduct, performs FS~\cite{fonti2017feature}. LASSO is a linear model formed by a vector of weights $\boldsymbol{\alpha}$ that can be used in classification and prediction tasks. Suppose for simplicity that we focus first on the static variables $\mathbf{z}_i\in\mathbb{R}^G$ and that all the entries of $\mathbf{z}_i$ are numerical. Then, the LASSO aims at finding the optimal value of $\boldsymbol{\alpha}\in\mathbb{R}^G$ that minimizes the cost
\begin{equation}
    \min _{\boldsymbol{\alpha} \in \mathbb{R}^{G}} \frac{1}{2}\sum_{i=1}^I \left(y_i - \mathbf{z}_i^\top\bbalpha \right)^{2}+\lambda\sum_{g=1}^G |\alpha_g|,    
    \label{EqLASSO_scalar}
\end{equation}
where $\|\bbalpha\|_{1}=\sum_{d=1}^D |\alpha_d|$ is the $ \ell_{1}$ norm of $\bbalpha$, and $\lambda > 0$ is a regularization parameter. Note that the cost function combines at the same time a data-fitting term with a regularizer term that penalizes the coefficients, shrinking some of them to zero. Trying to minimize the cost function, LASSO will automatically select the most informative features, discarding the useless or redundant ones. Therefore, the idea of using LASSO for FS purposes is to fit the model and then consider only the features $g$ with a coefficient $\alpha_g$ different from 0.

The LASSO method can be applied for the static data; however, since we are also dealing with MTS, we need to implement a modification of LASSO that can deal with matrices,  referred as \emph{Group LASSO}~\cite{chesneau2008some}. The main idea behind Group LASSO is to split the input feature into different groups and then either consider as relevant the entire group or eliminate all the variables within the group. We start by defining $\bbalpha^d=[\alpha_1^d,\alpha_2^d,...,\alpha_T^d]$, whose entries are associated with the $T$ samples recorded for feature $d$. Since we have $D$ vectors, the total number of coefficients to learn is $DT$. The optimal regularized regressor for the MTS features is obtained as the solution to
\begin{equation}\label{EqLASSO_vector}
\min_{\{\bbalpha^d\in\reals^T\}_{d=1}^D} \frac{1}{2}\sum_{i=1}^I \left(y_i- \sum_{d=1}^D  (\ubx_i^d)^\top \bbalpha^d \right)^{2}+\lambda   \sum_{d=1}^D\|\boldsymbol{\alpha}^d\|_2,
\end{equation}
where we recall that $\ubx_i^d$ is the vector collecting the entries of the $d$-th row of $\barbX_i$, and $\|\boldsymbol{\alpha}^d\|_2=((\alpha_1^d)^2+...+(\alpha_W^d)^2)^{1/2}\geq 0$ is the $\ell_{2}$ norm of $\bbalpha^d$. The above optimization resembles that in Eq.~\eqref{EqLASSO_scalar}, but accounting for the multidimensional nature of the input and replacing $|\alpha_d|$ with $\|\bbalpha^d\|_2$. This way, if the optimal solution sets $\bbalpha_*^d=[0,0,...,0]^\top$, then the $d$-th row of matrices $\{\barbX_i\}_{i=1}^I$ is not selected~\cite{chesneau2008some}. Clearly, upon using a binary cross entropy cost and a logistic regressor, the formulations in \eqref{EqLASSO_scalar} and \eqref{EqLASSO_vector} can be adapted to deal with classification problems.

\vspace{1mm}
\noindent\textbf{Permutation Feature Importance:}~
PFI is an FS method that leverages an already trained (black-box) architecture to identify the features that are more relevant for the output generated by the architecture. PFI attempts to emulate the (greedy) FS process that trains the architecture with all possible combinations of features while maintaining a commitment to computational cost. To reduce the computational cost, PFI does not train the model multiple times but evaluates the performance loss by perturbing each of the features of the input data. Specifically, to asses how important the input feature $d$ is, the PFI scheme replaces the value of feature $d$ with another input feature (say the $d'$-th one) and evaluates the performance loss associated with that permutation. PFI was originally proposed in~\cite{breiman2001random, breiman2001statistical} for \textit{random-forest} classifiers, and it has been successfully generalized to other setups~\cite{huang2016permutation,fisher2019all}. In our work, we have used the PFI method over several trained architectures, as described next. 

The first step of the PFI method is to train the ML model at hand and evaluate the classification performance of the trained classifier (according to a prespecified figure of merit) using the samples in the validation set. The next step is to select one of the features (say the $d$-th one), perturb each sample in the validation set by permuting the value of the $d$-th feature with one feature chosen uniformly at random, and keep all other features $d'\neq d$ as in the original validation set. By permuting the values of the feature under study, we do not change the marginal distribution of the feature, but we do ``break'' the relationships learned by the black-box model. After permuting feature $d$, we evaluate the same figure of merit using the modified validation set and compare the new value with that obtained using the original validation set. The rationale is that permutation of relevant features will lead to large Accuracy losses~\cite{gomez2020selecting}. The whole process is repeated to quantify the loss associated with the permutation of each feature $d=1,..,D$, and then the $D'$ most relevant features are selected.

\subsection{Mechanisms for Interpretability}
\label{subsec:Interpretable_methods}

The interpretable DNN presented in this section grasps knowledge using as a baseline the black-box DNN introduced in Sec. \ref{sec:NN}. Beyond providing insights, interpretable methods can also be used for fairness, accountability, and responsibility~\cite{Arrieta2020}. We can differentiate between two different families of mechanisms for interpretability~\cite{Gunning2019}. The first family builds "white-box" DNN models that are interpretable by design~\cite{Rudin2019}.  This work implements two different attention mechanisms as "white-box" interpretable models.

The second family of interpretable techniques generates post-hoc extrinsic explanations based on previous black-box trained models, considering only the model output while disregarding the model's internal mechanisms~\cite{Guidotti2018}. We also design and implement two post-hoc schemes: Time Perturbation Importances (TPI) and Dynamic Mask (Dynamask).

\subsubsection{Attention Mechanisms}
As shown in~\cite{kaji2019attention, remy2017keras}, attention-based models generate useful results that provide insights into the behavior of the classifier at the level of the input variables. 
The attention mechanism was originally developed for machine translation models~\cite{bahdanau2014neural}, although it has been successfully applied to very different problems, like medical computer vision tasks~\cite{sinha2020multi}, ECG analysis~\cite{zhang2019application}, and blood pressure response~\cite{gandin2021interpretability}.  The mechanism is capable of recognizing interactions between different time-slots and features, identifying how some time-slots influence others.

As in~\cite{kaji2019attention}, our attention mechanism operates at the input-variable level by using a dense layer with a softmax activation function. More specifically, for each patient $i$, an attention matrix $\mathbf{A}_{i}\in \mathbb{R}^{D\times T_i}$ is generated by postulating and learning an MLP mapping that takes $\textbf{X}_i \in \mathbb{R}^{D\times T_i}$ as input and yields $\mathbf{A}_{i}$ as output. 
The attention matrix $\mathbf{A}_{i}$ is used then for weighting the original input $\textbf{X}_i$ performing a Hadamard product, generating the weighted (attention modulated) input $\tilde{\textbf{X}}_i$. The role of the Hadamard product and the learnable matrix $\mathbf{A}_{i}$ is to endow the architecture with the ability to focus on the specific feature-time instant pairs that are more relevant for patient $i$. 

In this work, we propose two modifications relative to the attention mechanism in~\cite{kaji2019attention}.
Our first approach, which we label as the Non-Linear Hadamard Attention (NLHA) model, implements the original idea in \cite{kaji2019attention} but replacing the MLP with a GRN. The second approach, which we label as Hadamard Attention Matrix (HAM) model, deviates a bit more from the attention mechanism in \cite{kaji2019attention}. In particular, in lieu of learning an attention matrix $\mathbf{A}_{i}$ for each of the patients $i=1,...,I$, HAM learns a single matrix 
$\ubA$
and then applies the (same) weighting matrix 
$\mathbf{A}_{i}=\ubA$ 
to all the MTS $\mathbf{X}_{i}$ with $i=1,...,I$.

To preserve the clinical interpretability of $\mathbf{A}_i$, we emphasize that our designs apply the attention matrix directly to the input data $\mathbf{X}_i$, before any transformation/embedding is applied to the input. As a result, the entries of $\mathbf{A}_i$ can be readily used to assess the global contribution of each of the  $(d,t)$ feature-time instant pairs to the classification architecture.

\subsubsection{Time Perturbation Importances}
\label{subsec:TPI}

TPI is an inspection method to identify the most relevant time-slots based on already trained models. TPI is very similar to PFI but modified to account for the fact that the information at hand is an MTS.  
More specifically, TPI analyzes the performance degradation (e.g., the loss in classification Accuracy) of a previously trained model when the information associated with a particular time instant is corrupted. The main difference relative to PFI is that, rather than permuting the information with that of a different time instant (which would break the temporal structure of the data), TPI perturbs the original data by adding white Gaussian noise. 

More precisely, the first slot of the TPI method is to train a particular model and evaluate a desired figure of merit on a set of samples in the validation set. Then, after selecting a particular time instant (say the $t$-th one), the information of all the patients and features of the validation set for instant $t$ is perturbed, while the information for the time instants $t'\neq t$ remains unchanged. Finally, we compare the figures of merit after perturbing each time-slots separately with the value obtained for the original unperturbed validation. The larger the degradation, the more important the time instant for the problem a hand.

\subsubsection{Dynamic Mask}

Dynamask is a perturbation-based post-hoc method for MTS processing architectures that leverages already trained black-box models to provide knowledge about the importance of entries (i.e., features and time instants) of the input MTS~\cite{crabbe2021explaining}. 
The concept of post-hoc masks is widely used in image classification~\cite{fong2017interpretable, fong2019understanding}, with the goal being highlighting/identifying the regions of the picture that are salient for the operation of a black-box classifier. These masks are acquired by applying space-aware perturbations to the pixels of the original picture according to the value of the surrounding pixels and considering the effect of such perturbations in the output of the classifier. 
Dynamask modifies this perturbation idea using dynamic (time-aware) perturbations to MTS. Concretely, a dynamic perturbation is introduced so that the value of a feature at a particular time instant is replaced by a smoothened (filtered) version that averages (weights) the value of this feature at previous time-slots.

To describe this process more formally, suppose that our classifier has already been trained, recall that the input to the classifier is the MTS matrix $\mathbf{X}_{i}\in\mathbb{R}^{D\times T_i}$, and suppose that the label generated by the classifier for such input is $\hat{y}_i$. Furthermore, let matrix $\mathbf{M}\in\mathbb{R}^{D\times T_i}$ denote the matrix mask where the saliency scores are collected. The mask ${\mathbf{M}}$, together with the \emph{perturbation operator} $\pi$, are then used to generate the perturbed input ${\mathbf{X}}_i^P$ as $\mathbf{X}_i^P=\pi(\mathbf{X}_{i},\mathbf{M})$. The perturbed ${\mathbf{X}}_i^P$ is then fed to the black-box classifier to produce a perturbed output $\hat{y}_i^P$. The perturbed output $\hat{y}_i^P$ is finally compared to the original prediction $\hat{y}_i$, and the error is backpropagated to adapt the saliency scores collected in $ \mathbf{M}$. Repeating this process over many inputs and epochs, the values of the $\mathbf{M}$ are learned. 

While Dynamask can work with a wide range of perturbation (time-averaging) operators, to facilitate interpretation, we have implemented a simple moving average, so that the value of entry $(d,t)$ of the matrix $\mathbf{X}_i^P=\pi(\mathbf{X}_{i},\mathbf{M})$ is simply 
\begin{equation}\label{E:perturbation_operator_dynamask}
m^{(t, d)} x_i^{(t,d)}+(1-m^{(t, d)}) \mu_i^{(t, d)},~\text{with}~\mu_i^{(t, d)}=\frac{1}{W+1} \sum_{t^{\prime}=t-W}^{t} x_i^{(t^{\prime},d)},    
\end{equation}
and $W$ is the width of the window. Clearly, for the initial $t\leq W$ time instants, the definition of $\mu_i^{(t,d)}$ needs to be modified to account for the fact that less than $W$ input values are available. Under the perturbation operator in \eqref{E:perturbation_operator_dynamask}, it follows that values of $m^{(t, d)}$ close to one imply that the current value of $x_i^{(t,d)}$ is deemed important for the classifier, while values of $m^{(t, d)}$ close to zero imply that the current value of the input can be replaced by an average of the previous time-slots. Motivated by this and to foster interpretability, we augment the training cost of the Dynamask architecture with a penalty (regularizer) that promotes the values of the mask to be sparse and bounded by one (see~\cite{crabbe2021explaining} for additional details on the design of sparse masks).

\section{Dataset and Pre-Processing}
\label{sec:Database}

In this work, we have collected clinical data from the University Hospital of Fuenlabrada (UHF) in Spain over 17 years, from the beginning of January 2004 to the end of February 2020. A careful process of anonymization has been performed to preserve the identity of the patients. 
Data is associated with 2,784 patients, with a total of 3,158 ICU stays. The reason for having more stays than patients is that the same patient may have been admitted to the ICU more than once. Nonetheless, to use all available information, we consider these additional stays as new patients and work with $I=3,158$ samples.

To identify the presence of a multidrug-resistant germ, the microbiology laboratory staff performs two sequential procedures: first, microbiological culture and then the corresponding antimicrobial susceptibility test (named antibiogram). The culture is the process of isolating the germ that produces the infection, while the antibiogram is the procedure to test if the isolated germ is resistant to a set of antibiotics. Both processes together usually require at least 48 hours. % \textcolor{red}{Revisar - más elaboración lo de las 48 horas, separando los procesos}.
From a clinical viewpoint, we have limited this research to the first culture identified as multiresistant. For the 3,158 ICU patients, just in 605 cases, there was an AMR culture and, as a result, we dealt with a classification task where classes were significantly imbalanced.

Regarding the modeling of our MTS, we identify the first time slot ($t=0$) as the day the patient is admitted to the ICU. On the other hand, the last day $t=T_i$ depends on the type of stay. If the patient is non-AMR (i.e., if $y_i=0$), $T_i$ identifies the time slot the patient left the ICU. If the patient is AMR (i.e., if $y_i=1$), $T_i$ identifies the slot the culture was identified as AMR. Clearly, the length of the the considered MTS can be quite variable. To partially address this, following the literature and the clinical expertise, we 
implement a temporal windowing of 14 days. Mathematically, if the sampling period is 24 hours, this implies that for a given patient, say the $i$-th one, we have that $T_i<14$, then we use as input the original MTS $\mathbf{X}_i$. In contrast, if $T_i\geq 14$, we use as input the first $14$ columns of $\mathbf{X}_i$.
Previous studies have shown that models using relatively long windowing (within a reasonable length) and MTS of irregular length achieve better performance predicting the AMR onset than those that consider shorter windows or impute missing values to guarantee that all MTS have the same length~\cite{martinez2022interpretable}. 
The are several reasons to set the window duration to 14 days. For example, the first two weeks in the ICU are critical for the emergence of AMR germs~\cite{hinman1992meeting}. Also, when a patient is identified as infected by an AMR germ, the UHF clinical team quarantines the infected patient in the ICU for 14 days, which is a standard in the clinical setting~\cite{thombley2010menu}.

The data set contains both static variables and MTS. We consider these data to model both the initial patient's health status and the corresponding temporal evolution.  The static variables refer to demographical data and data associated with the patient's health status at the moment of the ICU admission. According to the clinical knowledge of the ICU clinical team at the UHF, we consider the following eight static variables: age, gender, the year when the patient was admitted, the month when the patient was admitted, the reason for the ICU admission, the clinical unit from which the patient comes (Origin), the category of the patient, and the SAPS-3 score (see~\cite{martinez2019machine} for more information about the static features). 

We now shift our attention to the MTS, which model the patient's health status evolution. Each variable (feature) registered in the MTS corresponds to one of the following three groups: features related to the patient's cultures, features associated with the patient's treatments, and features modeling the ICU occupation and the treatments followed by the rest of the patients who simultaneously stay in the ICU. 

The features linked to the patient's cultures allow us to identify the time-slots in which a germ has been found in any culture. Although cultures can identify multiple germs, only six of them are capable of becoming multidrug resistant: \textit{Pseudomonas}, \textit{Stenotrophomonas}, \textit{Acinetobacter}, \textit{Enterobacter}, \textit{Staphylococcus Aureus}, and \textit{Enterococcus}. 
For this reason, we have created MTS containing six variables (one per germ), counting the number of cultures per time-slot in which these germs have emerged. 
In the following sections, we use the name of the germ followed by the $pc$ (previous cultures) subscript to denote the features presented in this paragraph. For example, the feature modeling the emergence of \textit{Pseudomonas} is denoted as $Pseudomona_{pc}$.
We complete the MTS with an additional variable named $Others_{pc}$, which counts the number of germs that, do not belong to the set of six resistant germs, that were found in previous cultures. The variable  $Others_{pc}$ has been included to consider the possibility that some germs may be precursors to the appearance of multi-resistant germs. Note that, since we are trying to predict the onset of the first AMR infection per patient,  the germs modeled in the six variables in previous cultures were not multidrug-resistant. 

Regarding the features associated with the patient's treatment under study, we consider: (i) the mechanical ventilation variable,  denoting whether the patient has been connected (or not) to a breathing machine; and (ii) the families of the antibiotics taken by the patient during the ICU stay. The considered families are: A\-mi\-no\-gly\-co\-si\-des (AMG), An\-ti\-fun\-gals (ATF), Car\-ba\-pe\-ne\-mes (CAR), 1st generation Ce\-pha\-los\-porins (CF1), 2nd generation Ce\-pha\-los\-po\-rins (CF2), 3rd ge\-ne\-ra\-ti\-on Ce\-pha\-los\-po\-rins (CF3), 4th generation Ce\-pha\-los\-po\-rins (CF4), unclassified antibiotics (Others), Gly\-cy\-cli\-nes (GCC), Gly\-co\-pep\-ti\-des (GLI), Lin\-co\-sa\-mi\-des (LIN), Li\-po\-pep\-ti\-des (LIP), Ma\-cro\-li\-des (MAC), 
Mo\-no\-bac\-ta\-mas (MON), 
Ni\-troi\-mi\-da\-zo\-lics (NTI), Mis\-ce\-lla\-ne\-ous (OTR), Oxa\-zo\-li\-di\-no\-nes \hspace{0.5mm} (OXA), Broad-Spec\-trum  Pe\-ni\-ci\-llins \hspace{0.5mm}(PAP), \hspace{0.3mm} Pe\-ni\-ci\-llins \hspace{0.3mm}(PEN), \hspace{0.3mm} Po\-ly\-pep\-ti\-des \hspace{0.3mm}(POL), \hspace{0.3mm} Qui\-no\-lo\-nes \hspace{0.3mm} (QUI), Sul\-fa\-mi\-des (SUL), and Te\-tra\-cy\-cli\-nes (TTC). We also use the feature ``Others'' to identify any other family of antibiotics not belonging to the previous list.  
Thus, for a particular patient (say the $i$-th one), the feature associated with each treatment (say the $d$-th one) is a sequence of binary variables $\ubx_i^d\in \{0,1\}^{D\times T_i}$ indicating whether the patient has received (or not) the treatment during each of the $T_i$ time-slots (24-hour periods)  the patient stayed in the ICU. 

As for the last group of the MTS, we represent both the ICU occupation and a summary of the antimicrobials taken by the remainder of the ICU patients (neighbors) during the same time-slots considered for the patient under study. Therefore, we have modeled 25 extra numeric features: the number of neighbors of the patient under study,  the number of patients identified with AMR bacteria (\# of AMR neighbors), and the number of neighbors taking each of the 23 antibiotic families previously indicated. We use the subscript $n$ in the name of the variable to denote features referring to the neighbors of the patient under study. This way, and as an example, the variable $CAR$ is the feature indicating if the patient under study took that drug, while $CAR_n$ is the feature counting the number of neighbors who also took that drug.

\section{Results for Early Prediction and Interpretability  of AMR Using Multimodal Data}
\label{sec:Experiments}

We first explain in this section the experimental setup. 
Secondly, we introduce the figures of merit considered for evaluating the models' performance. After that, we present a set of experiments considering all the available features to identify the early prediction of AMR and analyze the obtained performance. Then, since we are tackling a very complex problem with a limited number of samples and a considerable number of features, we study the effect of applying a knowledge extraction process. This process is composed of an FS technique followed by interpretable methods. Finally, we present and discuss the prediction performance of models trained after studying the knowledge extraction results.

\subsection{Experimental Setup and Parameter Tuning}
\label{Sec:ExperimentalSetup}

It is expected that models trained using ML techniques provide good generalization capabilities, i.e., that they provide reasonable outputs when considering samples not used during the model design~\cite{doshi2018considerations}. To estimate and compare the generalization capabilities of different models, it is necessary to separate the data set into two independent subsets: the training set and the test set. The training set is used to construct the model through a learning process, and the test set is used to evaluate the performance of the built model. According to the literature, we decided to assign 80\%  of the samples to the training set and the remaining 20\% of the samples to the test set. To avoid bias considering just one random split, it is usual to repeat the train-test split several times, creating different models and evaluating each of them with the corresponding test set. In this work, we have performed three random splits of the train-test sets, always providing performance on the test sets.

We followed a 5-fold cross-validation approach in the training set to select the hyperparameters minimizing the Balanced Binary Cross-Entropy (BBCE) cost function using the optimization algorithm Adam~\cite{Adam_15, stone1978cross}. 
The hyperparameters associated with the  MLP, GRU, JHF, and FHSI network architectures are the learning rate, explored considering the values $\{$0.0001, 0.001, 0.01, 0.1$\}$, the dropout rate $\{$0.0, 0.1, 0.2, 0.3$\}$ and the number of neurons in the hidden layers $\{$3, 5, 8, 10, 15, 20, 25, 30, 35, 40, 50$\}$. We have chosen the widely-used Leaky Rectified Linear Unit (ReLU) as non-linear activation function~\cite{NairRELU2010}.
To avoid overfitting, we have applied an early-stopping technique~\cite{yao2007early}. At every epoch, the early-stopping approach evaluates the cost in the validation set and stops the training when the cost increases or stagnates.
Before training, each feature was normalized to have zero mean and standard deviation one~\cite{Duda_01}.

We have used a cost-sensitive learning strategy to deal with imbalanced classes in this work. The asymmetrical loss function used is the BBCE function, a widely used modification of the binary cross-entropy cost function~\cite{aurelio2019learning}. The BBCE loss function considers a weighting factor $\beta \in (0,1)$ to modify the penalty of failing on the minority class prediction. More specifically, the BBCE function is defined as 
\begin{equation}
\label{Eq:BBCE}
 {\cal L}_{BBCE} = -\frac{1}{I'}\sum_{i=1}^{I'}\left( \beta y_{i} \log (\hat{y}_{i})+ (1 - \beta) (1-y_{i})\log (1-\hat{y}_{i})\right) 
\end{equation}
where $I'$ is the number of samples in the training set. Following the recommendations in the technical literature~\cite{aurelio2019learning} and our previous work~\cite{martinez2022interpretable}, we have set the  BBCE weight as the number of samples of the majority class divided by the total number of samples. This way, the parameter $\beta$ is greater than 0.5 and the design penalizes failing in the minority class.

All our architectures (except the MLP) are intended to return a time series as output. However, the work undertaken is a binary classification problem with a non-vector label.
Because GRU-based architectures assume that information from previous time-slots is contained in the memory of the architecture, we have decided to use the value returned for the last time-slot as output.

\subsection{Performance Evaluation}
\label{sec:val}
There are several figures of merit to evaluate the ability of a model to make correct predictions, being Accuracy the most used in the literature. It measures the ratio between the correctly classified samples and the total number of samples under consideration~\cite{international1994accuracy}. 
However, using the classification Accuracy can overestimate the model performance due to the class imbalance.  For this purpose, we have considered in this work other figures of merit such as Specificity, Sensitivity, and the Receiver Operating Characteristic Area Under the Curve (ROC AUC)~\cite{davis2006relationship}. The Sensitivity indicates the ratio of AMR samples classified as AMR; Specificity considers the ratio of non-AMR samples classified correctly by the model as non-AMR. Finally, the ROC AUC measures the overall performance of a binary classifier~\cite{mcclish1989analyzing}, giving insights into the interdependency between Specificity and Sensitivity.

\subsection{Results Considering All Features}
\label{sec:results_models_no_FS}

The main goal of this work is to predict the early emergence of AMR with multimodal data recorded in the EHR. We will first compare non-multimodal (MLP and GRU) and multimodal (JHF, FHSI, LFCO) data-driven models using all the features. Table~\ref{Table:results_noFS} shows the mean and the standard deviation computed using the three test splits in terms of Accuracy, Specificity, Sensitivity, and ROC AUC. To keep fairness with all methods, the same three test sets have been considered in all the experimental work. 

The MLP yields the worst results (62.29 ROC AUC), probably because it does not consider data about the patient's evolution. It is the only architecture considering just the static variables. When focusing on architectures using MTS, both GRU (75.50 ROC AUC) and multimodal models (76.33 $\pm$ 0.27 ROC AUC) improve the results provided by the MLP. Note that the GRU and the multimodal architectures provide pretty similar results (it should be pointed out that the multimodal architectures use both MTS and static variables).

Following the approach taken in~\cite{martinez2022interpretable}, we will perform different experiments 
based on FS methods and interpretable mechanisms
to potentially gain knowledge and train models, improving the performance presented in Sec.~\ref{sec:val}.

\begin{table*}[t!]
    \centering
    {
        \begin{tabular}{|c | c | c| c| c|}
        \hline
        \textbf{Method} & \textbf{Accuracy} & \textbf{Specificity} & \textbf{Sensitivity} &  \textbf{ROC AUC}   \\
        \hline
        MLP &  58.60 $\pm$ 0.52 & 58.62 $\pm$ 0.48 & 58.37 $\pm$ 4.64 & 62.29 $\pm$ 2.34\\
        GRU &  63.19 $\pm$ 2.47 & 59.91 $\pm$ 4.17 & 77.83 $\pm$ 5.83 & 75.50 $\pm$ 0.36\\
        FHSI & 62.76 $\pm$ 3.25 & 59.17 $\pm$ 4.45 & {78.98 $\pm$ 3.56} & \textbf{76.74 $\pm$ 1.36}\\
        JHF &  65.14 $\pm$ 1.55 & 62.58 $\pm$ 1.29 & 76.55 $\pm$ 1.80 & 76.20 $\pm$ 1.17\\
        LFLR & \textbf{67.25 $\pm$ 2.29} & \textbf{65.90 $\pm$ 3.56} & 73.75 $\pm$ 3.76 & 76.21 $\pm$ 1.31\\
        LFCO & 60.92 $\pm$ 3.14 & 56.39 $\pm$ 4.38 & \textbf{81.38 $\pm$ 3.53} & 76.18 $\pm$ 1.31\\
        \hline
        \end{tabular}
    }
    \caption{Mean $\pm$ standard deviation of the performance (Accuracy, Specificity, Sensitivity, and ROC AUC) on three test partitions when training the classification architectures considering all the features. The highest performance for each figure of merit is in bold.}
    \label{Table:results_noFS}
\end{table*}

\subsection{FS and Interpretable Mechanisms for Knowledge Extraction}

To improve the results and potentially gain knowledge about the inherent mechanism of the AMR onset, we propose to perform different FS procedures as well as analyze several interpretable mechanisms.

\begin{figure*}[ht!]
    \centering
	\includegraphics[width=\textwidth]{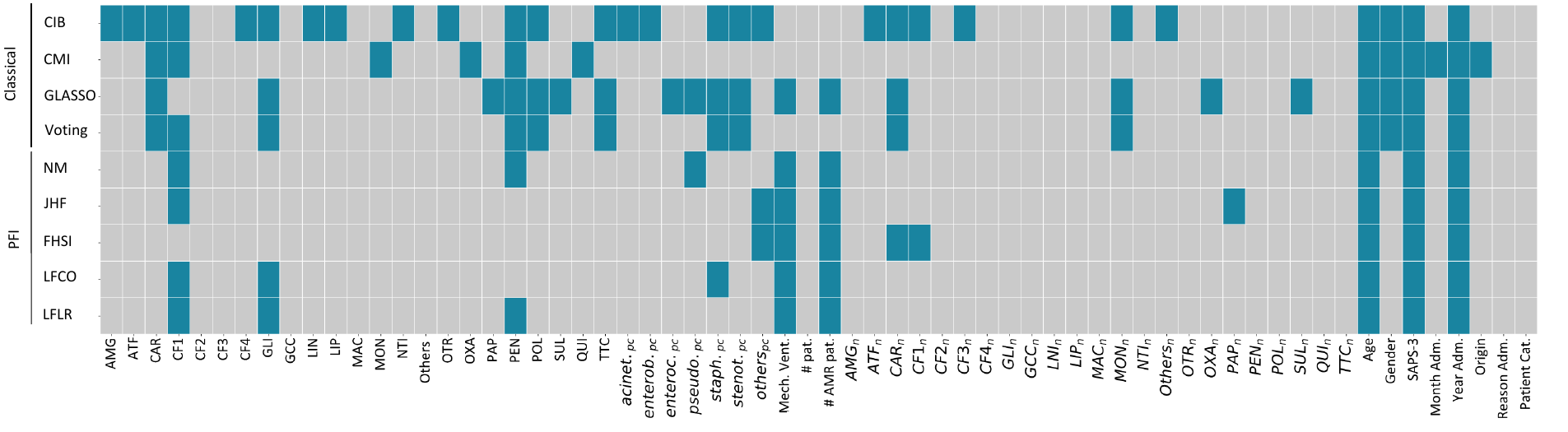}
	\caption{
	Matrix of features (in columns) and FS approaches (in rows, organized as classical and PFI techniques). The blue cells (darker ones) represent the selected features. Note that the NM results consider two different models: a MLP for dealing with static data and an RNN when considering MTS.}

	\label{fig:heatmapFS}
\end{figure*}

Firstly, we pay attention to the FS process. Figure~\ref{fig:heatmapFS} shows a matrix where variables are in columns and techniques presented in Sec.~\ref{subsec:FeatureSelection} are in rows. When the cell is marked in blue (darker ones), it indicates that the corresponding method has selected that feature. Both classical and PFI techniques have been analyzed.  The upper part of the matrix shows the results of the classical FS methods (CIB, CMI, and GLASSO), while the lower part is associated with the application of PFI on each of the models presented in Sec.~\ref{sec:results_models_no_FS}. 
We also implemented a majority voting scheme among the three classical FS methods. Thus, according to the voting scheme, the $d$-th feature is selected by at least two of the classical schemes. 

Paying now our attention to the PFI results, the five different implementations select: among the static variables,  the age of the patient, the SAPS-3 score, and the year of admission; and the mechanical ventilation and the number of AMR neighbors among the MTS variables. Regarding the antibiotics administered during the patient stay, note that CF1 and PEN are selected by three of the five PFI implementations. 

The classical FS methods select a wider range of features, especially in the MTS case. Note that the classical FS methods do not agree as much with the variables they select as the FSI methods, with a considerable number of features being selected by only one of the classical FS methods.
All the classical FS methods select CAR, and PEN, while  the PFI implementations previously selected the PEN antibiotic family. In the static case, the age, gender, SAPS-3, and the year of the admissions are also selected by all the methods. 

According to the clinical knowledge of the UHF staff, the selection of features such as SAPS-3 score, mechanical ventilation, and the number of AMR neighbors is consistent with the clinical literature.
Once the FS approaches have been applied, we will study the selected features using the interpretable mechanisms presented in Sec.~\ref{subsec:Interpretable_methods}.

\begin{figure*}[ht!]
    \centering
    \includegraphics[width=\textwidth]{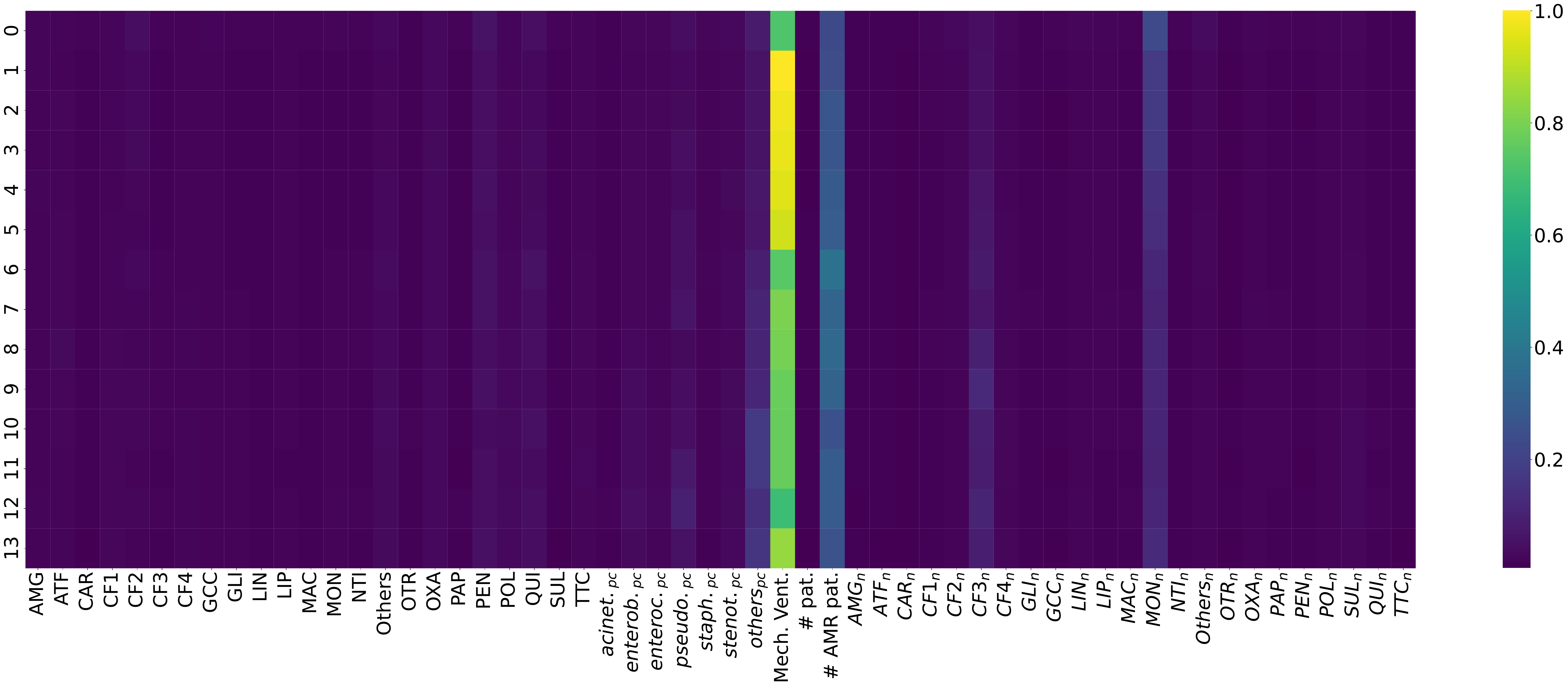}
	\caption{
	Importance score heatmap using all the features as input representing the average of the $A_{i}$ matrices corresponding to the NLHA model. Columns represent features, while rows show time-slots of the MTS under study (`0' refers to the day of the ICU admission).
	}
	\label{fig:attention_scores_NLHA}
\end{figure*}

Following the FS results, we analyze the scores obtained when applying the implemented attention mechanisms (NLHA and HAM). Firstly, we show in Figure~\ref{fig:attention_scores_NLHA} a heatmap representing the attention scores obtained when applying the NLHA mechanism using the FHSI model because, as demonstrated next, it yields the best performance. The columns of the heatmap represent features, while rows show time-slots of the MTS under study (`0' refers to the day of the ICU admission). Since the heatmap represents importance scores for both features and time-slots, only the MTS are represented and the static variables are excluded from this figure. 

Since NLHA generates an attention matrix $A_{i}$ for each sample, Figure~\ref{fig:attention_scores_NLHA} represents the average across all the attention matrices. Note that mechanical ventilation is the variable with the highest importance score, followed by the number of AMR neighbors of the patient. These results are in accordance with previous results of the PFI techniques. The heatmap also shows higher scores in the mechanical ventilation feature during the first days of the patient's stay.

\begin{figure*}[ht!]
    \centering
    \includegraphics[width=\textwidth]{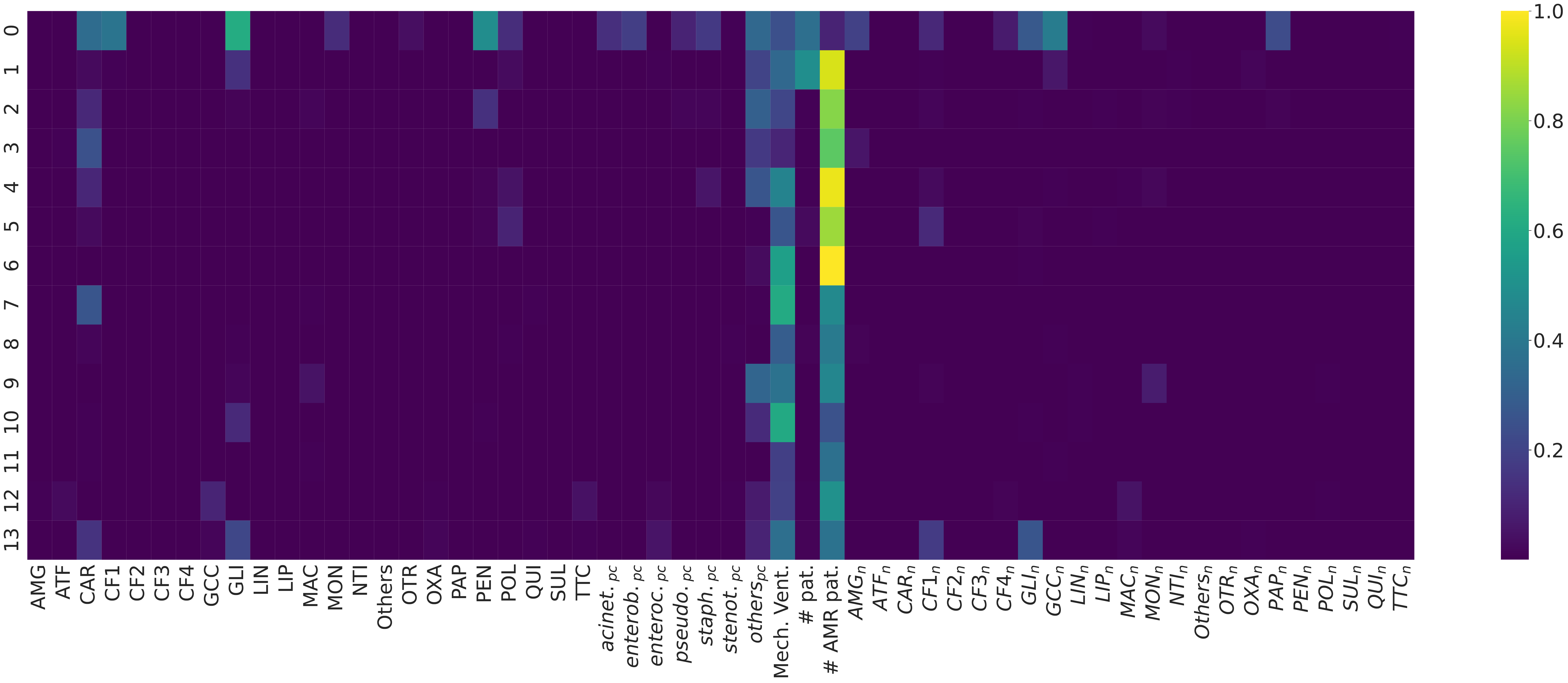}
	\caption{
	Importance score heatmap representing the matrix $\ubA$ of the HAM model
	when using all the MTS variables. Columns represent features, while rows show time-slots of the MTS under study (`0' refers to the day of the ICU admission).}
	\label{fig:attention_scores_HAM}
\end{figure*}

Once we have analyzed the importance scores provided by the NLHA architecture, we will now proceed with the analysis of another attention mechanism presented, HAM. The scores of attention corresponding to the matrix $\ubA$ of the HAM architecture using the FHSI black-box model are presented in Figure~\ref{fig:attention_scores_HAM}.
The representation in Figure~\ref{fig:attention_scores_HAM} is the same as in Figure~\ref{fig:attention_scores_NLHA}, the columns represent features, while rows show time-slots of the MTS under study (`0' refers to the day of the ICU admission). The importance of mechanical ventilation and the number of AMR neighbors is also evidenced here, with the early days of the patients' stay identified again as relevant. Some antibiotics such as CAR, GLI, or PEN also have high scores on the first day of the patient's stay.

\begin{figure*}[ht!]
    \centering
	\includegraphics[width=\textwidth]{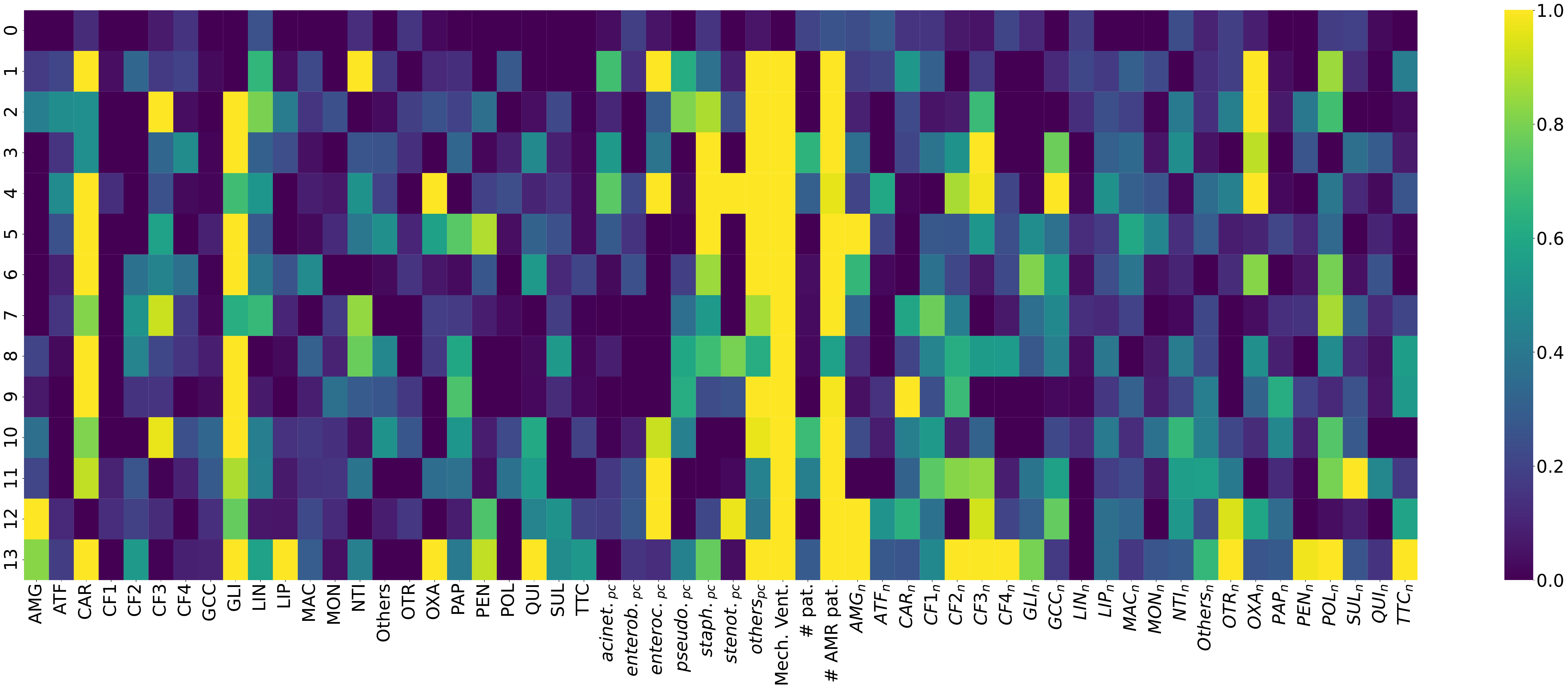}
	\caption{Heatmap of importance scores when using the Dynamask model over an already trained FHSI model using all the features. Columns represent features, while rows show time-slots of the MTS under study (`0' refers to the day of the ICU admission).}
	\label{fig:dynamask_scores}
\end{figure*}

Figure~\ref{fig:dynamask_scores} shows the scores after applying the Dynamask me\-cha\-nism using the FHSI model as black-box model. We have used the FHSI model because its ROC AUC is slightly better than the one yielded by the other models. Columns in Figure~\ref{fig:dynamask_scores}  represent features, while rows indicate time-slots of the MTS under study (`0' refers to the day of the ICU admission). As in  Figure~\ref{fig:attention_scores_NLHA} and Figure~\ref{fig:attention_scores_HAM}, the heatmap showed in Figure~\ref{fig:dynamask_scores} only shows MTS, since represents importance scores for features and time-slots. The importance of mechanical ventilation and the number of AMR neighbors is also illustrated in Figure~\ref{fig:dynamask_scores}. Recall that those features were also ranked with high scores in Figure~\ref{fig:attention_scores_NLHA} and Figure~\ref{fig:attention_scores_HAM}.
The Dynamask mechanism also assigns high scores to features such as the CAR antibiotic family, the results of previous cultures with non-AMR germs, and the number of neighbors of the patient.

The LFCO model presented in Sec.~\ref{subsec:late_fusion_method} can also help us to gain knowledge about the task to solve. The weight $w_{MLP}$ and $w_{GRU}$ showcase the degree of importance of the static and  MTS variables for the particular task. High values (greater than 0.5) of $w_{MLP}$ indicate that static variables are more important, while low values of $w_{MLP}$ suggest that MTS are more relevant for the prediction (recall that, as explained in Sec.~\ref{subsec:late_fusion_method}, the only constraint in the LFCO is that $w_{GRU}~+~w_{MLP}~=~1$). Our experiments show that the mean value of $w_{MLP}$, computed when considering the three training partitions, is 0.34 (standard deviation of 0.03). Therefore, we can conclude that considering the LFCO scores, the MTS are more important than the static variables. This statement is reinforced if we compare the MLP and GRU results in the previous section (see Table~\ref{Table:results_noFS}).

Owing to the temporal dimension of part of the considered data set, a study of the most relevant time-slots of the patient's stay has been performed. This analysis could reveal interesting information from a clinical viewpoint, allowing the physicians to be extremely vigilant on certain days to avoid the emergence of AMR germs.

We have performed the time-slots analysis using the TPI approach over the pre-trained FHSI model presented in Sec.~\ref{subsec:TPI}. TPI results indicate that the first (0.26), second (0.91), third (1.00), fourth (0.38), eleventh (0.29), and fourteenth (0.41) time-slots are those with the highest scores. The rest of the time-slots have obtained lower scores, with a mean value of 0.14. This suggests that: i) the first days of the stay are the most relevant for the prediction (this is expected, especially because there is a large number of patients who develop the resistance in the first 72 hours); and ii) for the patients whose MTS is windowed, the last day is important for the prediction. While the second point could indicate that longer windows should be considered, we ran an exploratory analysis and concluded that this was not the case.

\subsection{Results Considering the Selected Features}

To evaluate the benefits of designing a model for AMR prediction by following  FS strategies, we present in Table~\ref{Table:results_FS} the average and standard deviation on Accuracy, Specificity, Sensitivity, and ROC AUC when considering the same architectures as in Sec.~\ref{sec:results_models_no_FS}. Table~\ref{Table:results_FS} is organized into two groups of FS strategies (see column Data Source). In the first group, the classical FS strategies have been considered to train the models with the set of features selected by CIB, CMI, GLASSO, and the corresponding voting procedure. 
The second group refers to sets of features selected by the PFI techniques. Although experiments were carried out with more features,  the best results were obtained using 3, 4, and 5 MTS. Therefore, Table~\ref{Table:results_FS} shows the results in two scenarios: first using only the 3, 4, and 5 MTS selected by each approach, and then considering also the 3 static features (the age of the patient, the SAPS-3 score, and the year of admission). 
 
Some conclusions can be drawn from the table. First,  focusing on the FS methods, better performance is provided by the PFI strategy: the average ROC AUC value is 78.77 against 70.46 when considering the results of the 18 PFI approaches and the 24 classical FS methods, respectively. Among the classical FS techniques, CMI obtains the worst performance (64.22 ROC AUC in mean), followed by the voting strategy (64.38 ROC AUC in mean) and CIB  (average value of 76.30 for the ROC AUC). Finally, the best classical FS method is GLASSO, yielding an average value of 76.98 for the ROC AUC.

Regarding the considered architectures, MLP is the one providing the worst results. Note that the MLP already yielded the worst results in Sec.~\ref{sec:results_models_no_FS},  possibly because the use of static variables is quite limited to solving the task. Comparing the performance of the GRU approach (which only uses MTS, providing an average ROC AUC of 74.60) with that of the set of multimodal architectures (average ROC AUC of 76.85), it is noticeable the benefits of the multimodal models. Focusing now on the multimodal architectures, the best performance is yielded by the FHSI model trained with 3 MTS and 3 static features, both in terms of Accuracy (73.89 $\pm$ 3.55), Specificity (72.63 $\pm$ 5.43), and ROC AUC (84.33 $\pm$ 1.38).

\begin{table*}[t!]
    \centering
    {
        \begin{tabular}{|c |  c | c | c | c| c| c|}
        \hline
        
        \textbf{Data Source} &  \textbf{Method} & \textbf{Features} & \textbf{Accuracy} & \textbf{Specificity} & \textbf{Sensitivity} &  \textbf{ROC AUC}   \\
        \hline
        \multirow{24}{*}{\textbf{Classical FS }} &
            \multirow{4}{*}{\textbf{MLP}} 
                &   \textbf{CIB features}  & 58.12 $\pm$ 4.46 & 58.95 $\pm$ 6.74 & 54.38 $\pm$ 6.58 & 61.92 $\pm$ 1.40\\
                & & \textbf{CMI features}  &  49.74 $\pm$ 9.55 & 45.14 $\pm$ 13.53 & 70.96 $\pm$ 10.19 & 62.23 $\pm$ 1.23\\
                & & \textbf{GLASSO features}  & 58.12 $\pm$ 4.46 & 58.95 $\pm$ 6.74 & 54.38 $\pm$ 6.58 & 61.92 $\pm$ 1.40\\
                & & \textbf{Voting features}  &  58.12 $\pm$ 4.46 & 58.95 $\pm$ 6.74 & 54.38 $\pm$ 6.58 & 61.92 $\pm$ 1.40\\
            \cline{2-7}

            & \multirow{4}{*}{\textbf{GRU}} 
                & \textbf{CIB features}  & 64.14 $\pm$ 3.71 & 59.96 $\pm$ 3.78 & \textbf{82.75 $\pm$ 3.31} & 78.83 $\pm$ 3.39\\
                & & \textbf{CMI features}  & 37.29 $\pm$ 4.12 & 27.07 $\pm$ 7.64 & 81.84 $\pm$ 8.02 & 60.22 $\pm$ 3.24\\
                & & \textbf{GLASSO features}  & 68.72 $\pm$ 1.65 & 67.31 $\pm$ 2.50 & 75.38 $\pm$ 3.93 & 78.80 $\pm$ 1.62\\
                & & \textbf{Voting features}  & 47.68 $\pm$ 2.98 & 42.85 $\pm$ 3.18 & 69.43 $\pm$ 0.28 & 60.99 $\pm$ 2.79\\
            \cline{2-7}

            & \multirow{4}{*}{\textbf{JHF}} 
                &   \textbf{CIB features}  & 68.57 $\pm$ 2.50 & 67.04 $\pm$ 3.92 & 75.64 $\pm$ 4.38 & 79.05 $\pm$ 1.39\\
                & & \textbf{CMI features}  & 58.02 $\pm$ 3.30 & 55.70 $\pm$ 5.13 & 67.78 $\pm$ 3.86 & 65.11 $\pm$ 1.64\\
                & & \textbf{GLASSO features}  & 69.41 $\pm$ 2.47 & 67.97 $\pm$ 3.25 & 76.21 $\pm$ 3.33 & 80.07 $\pm$ 2.30\\
                & & \textbf{Voting features}  & 58.23 $\pm$ 0.93 & 56.76 $\pm$ 1.61 & 64.76 $\pm$ 4.08 & 65.12 $\pm$ 3.06\\
            \cline{2-7}

            & \multirow{4}{*}{\textbf{FHSI}} 
                &   \textbf{CIB features}  & 71.52 $\pm$ 3.23 & 70.43 $\pm$ 4.68 & 76.82 $\pm$ 3.96 & 81.01 $\pm$ 0.21\\
                & & \textbf{CMI features}  & 56.80 $\pm$ 1.46 & 54.07 $\pm$ 1.89 & 68.79 $\pm$ 2.36 & 66.66 $\pm$ 1.27\\
                & & \textbf{GLASSO features}  & 68.83 $\pm$ 4.13 & 65.95 $\pm$ 5.73 & 82.46 $\pm$ 4.13 & 81.76 $\pm$ 2.43\\
                & & \textbf{Voting features}  & 62.82 $\pm$ 2.49 & 63.44 $\pm$ 3.31 & 60.23 $\pm$ 1.58 & 66.95 $\pm$ 1.62\\
            \cline{2-7}
                
            & \multirow{4}{*}{\textbf{LFLR}} 
                &   \textbf{CIB features}  &  70.04 $\pm$ 1.81 & 68.78 $\pm$ 2.02 & 75.73 $\pm$ 0.74 & 78.49 $\pm$ 1.82\\
                & & \textbf{CMI features}  & 54.22 $\pm$ 3.14 & 50.49 $\pm$ 4.53 & 71.44 $\pm$ 4.40 & 65.55 $\pm$ 0.25\\
                & & \textbf{GLASSO features}  & 69.99 $\pm$ 0.49 & 68.51 $\pm$ 1.00 & 76.89 $\pm$ 2.91 & 79.34 $\pm$ 0.42\\
                & & \textbf{Voting features}  & 54.96 $\pm$ 5.53 & 51.36 $\pm$ 6.60 & 71.02 $\pm$ 2.09 & 65.72 $\pm$ 3.23\\
            \cline{2-7}
            & \multirow{4}{*}{\textbf{LFCO}} 
                &   \textbf{CIB features}  & 67.62 $\pm$ 1.38 & 65.49 $\pm$ 1.91 & 77.37 $\pm$ 1.81 & 78.47 $\pm$ 1.23\\
                & & \textbf{CMI features}  & 50.21 $\pm$ 5.18 & 43.66 $\pm$ 7.57 & 79.81 $\pm$ 7.14 & 65.55 $\pm$ 0.34 \\
                & & \textbf{GLASSO features}  & 68.93 $\pm$ 2.76 & 67.19 $\pm$ 4.13 & 77.30 $\pm$ 4.57 & 79.99 $\pm$ 0.66\\
                & & \textbf{Voting features}  & 52.58 $\pm$ 5.75 & 48.85 $\pm$ 6.70 & 69.66 $\pm$ 2.33 & 65.57 $\pm$ 2.78\\
        \hline
        \multirow{18}{*}{\textbf{PFI}} &
            \multirow{3}{*}{\textbf{MLP}} 
                & \textbf{3 features}  & 46.04 $\pm$ 3.77 & 39.54 $\pm$ 6.97 & 74.17 $\pm$ 7.98 & 62.09 $\pm$ 1.07\\
                & & \textbf{4 features}  & 62.29 $\pm$ 0.97 & 64.00 $\pm$ 1.56 & 54.88 $\pm$ 1.57 & 62.16 $\pm$ 0.71\\
                & & \textbf{5 features}  & 52.85 $\pm$ 2.02 & 50.50 $\pm$ 2.68 & 63.33 $\pm$ 2.27 & 62.60 $\pm$ 1.19\\
        \cline{2-7}
            &\multirow{3}{*}{\textbf{GRU}} 
                & \textbf{3 MTS}  & 67.51 $\pm$ 3.03 & 64.47 $\pm$ 3.22 & 81.16 $\pm$ 1.37 & 81.85 $\pm$ 1.43\\
                & & \textbf{4 MTS}  & 68.78 $\pm$ 2.90 & 66.42 $\pm$ 3.66 & 79.62 $\pm$ 1.35 & 80.88 $\pm$ 1.90\\
                & & \textbf{5 MTS}  & 67.14 $\pm$ 2.57 & 64.09 $\pm$ 2.64 & 80.93 $\pm$ 3.16 & 80.68 $\pm$ 2.44\\
        \cline{2-7}
            & \multirow{3}{*}{\textbf{JHF}} 
                &  \textbf{3 MTS + 3 feat.} & 71.89 $\pm$ 1.74 & 70.02 $\pm$ 2.10 & 80.49 $\pm$ 6.32 & 82.94 $\pm$ 2.01\\
                & &  \textbf{4 MTS + 3 feat.} & 69.36 $\pm$ 1.90 & 67.18 $\pm$ 2.41 & 79.35 $\pm$ 1.81 & 81.61 $\pm$ 1.06\\
                & &  \textbf{5 MTS + 3 feat.} &  69.78 $\pm$ 2.45 & 68.61 $\pm$ 2.72 & 75.23 $\pm$ 4.34 & 80.97 $\pm$ 2.24\\
        \cline{2-7}
            & \multirow{3}{*}{\textbf{FHSI}} 
                & \textbf{3 MTS + 3 feat.} & \textbf{ 73.89 $\pm$ 3.55} & \textbf{72.63 $\pm$ 5.43} & 79.47 $\pm$ 5.62 & \textbf{84.33 $\pm$ 1.38}\\
                & & \textbf{4 MTS + 3 feat.} & 71.94 $\pm$ 3.03 & 69.49 $\pm$ 4.53 & {82.27 $\pm$ 5.49} & 83.48 $\pm$ 2.68\\
                & & \textbf{5 MTS + 3 feat.} & 71.84 $\pm$ 1.81 & 69.86 $\pm$ 3.48 & 80.01 $\pm$ 4.82 & 82.92 $\pm$ 2.08\\
        \cline{2-7}
            & \multirow{3}{*}{\textbf{LFLR}} 
                &  \textbf{3 MTS + 3 feat.} & 68.88 $\pm$ 2.87 & 66.82 $\pm$ 3.68 & 78.56 $\pm$ 4.25 & 81.83 $\pm$ 1.69\\
                & &  \textbf{4 MTS + 3 feat.} & 68.93 $\pm$ 1.55 & 66.60 $\pm$ 2.44 & 79.84 $\pm$ 4.17 & 82.07 $\pm$ 1.28\\
                & &  \textbf{5 MTS + 3 feat.} &  68.09 $\pm$ 1.20 & 66.06 $\pm$ 1.10 & 77.28 $\pm$ 4.02 & 81.32 $\pm$ 1.85\\
        \cline{2-7}
            & \multirow{3}{*}{\textbf{LFCO}} 
                &  \textbf{3 MTS + 3 feat.} & 69.78 $\pm$ 1.71 & 68.26 $\pm$ 2.13 & 76.88 $\pm$ 2.98 & 82.25 $\pm$ 1.37\\
                & &  \textbf{4 MTS + 3 feat.} & 69.41 $\pm$ 1.47 & 67.61 $\pm$ 1.52 & 77.65 $\pm$ 2.40 & 81.50 $\pm$ 1.45\\
                & &  \textbf{5 MTS + 3 feat.} &  69.25 $\pm$ 1.30 & 66.42 $\pm$ 1.42 & 81.72 $\pm$ 1.67 & 82.32 $\pm$ 1.04\\
        \hline
        \end{tabular}
    }
    \caption{Mean $\pm$ standard deviation values of four figures of merit (Accuracy, Specificity, Sensitivity, and ROC AUC) on three test partitions when training the classification architectures
    considering: classical-FS and PFI techniques (first column); MLP, GRU, JHF, FHSI, LFLR, and LFCO classifiers (second column); and different sets of features (determined by the approaches in the third column). All the multimodal techniques (JHF, FHSI, LFLR and LFCO) use the same static variables (age of the patient, SAPS-3 score, and year of admission). The highest performance for each figure of merit is in bold. 
}
    \label{Table:results_FS}
\end{table*}

\section{Discussion and Conclusions}
\label{sec:Conclusion}

    AMR is a serious clinical issue whose severity is growing due to the improper use of antibiotics~\cite{hsu2020covid}. The growth of the AMR could endanger the viability of healthcare systems, affecting the cost of treatments, patient comorbidities, and the waste of resources~\cite{tansarli2013impact}. This leads to a significant worldwide problem permeating all hospital services, especially the ICU, due to the fragile health status of patients staying in this unit.

    Longitudinal EHR records patient health data over time and has proven to be
    one of the most valuable data resources for clinical decision support.  Data registered in  MTS, in conjunction with data from different sources, can accurately represent the patient's health status. However, dealing with EHR data, especially with MTS, is challenging since a vast amount of heterogeneous data is recorded for each patient at different and irregular time intervals.

Recent studies have explored the use of EHR data and DNN models to predict the occurrence of AMR, showcasing their suitability to speed up hospital workflow, reducing  and saving costs~\cite{martinez2022interpretable, shichijo2017application, shahid2019applications}. The complexity of the data contained in the EHR makes simple DNNs models unable to identify and model the complex relationships between the different multimodal data. Throughout this work, a set of multimodal DNNs using MTS and static features have been developed. The results have demonstrated the ability of multimodal DNN approaches to properly model the complex relationships of different data sources, showing better performance than non-multimodal models. However, DNN models are considered black-box models, which challenges their use in the clinical setting~\cite{london2019artificial}, since they often lack interpretability to understand the hidden patterns and to support decision-making. 

The multimodal DNNs models proposed in this work are not only able to perform well in predicting AMR but also have interpretable mechanisms. More precisely, we have used several tools based on FS methods and interpretable mechanisms to extract new clinical insights. The findings provided in this work could be used to support clinical decisions in the ICU. For example, both the FS results and the interpretable mechanisms have shown mechanical ventilation and the \# of AMR neighbors as crucial features in AMR development. Other families of antibiotics such as CAR, CF1, GLI, and PEN have also been identified by some methods, proving relevant knowledge in the AMR emergence.  All these findings agree with the existing literature: antibiotic families previously commented on are widely used, and invasive procedures such as mechanical ventilation have been proved to cause infections and drug resistance~\cite{tissing1993risk}.

The methodology presented throughout this work could help to perform antibiotic treatments more intelligently and organize patients in the ICU space to reduce germ transmission. In addition, it could stop possible AMR germ outbreaks in the ICU.  Since EHR data can be extrapolated to any clinical problem, the proposed methodology paves the way for multimodal and explainable prediction support systems that may be used to solve other clinical problems, expanding the relevance and application of our findings.

We close the manuscript with the identification of several future research directions. From a clinical point of view, we are looking at the incorporation of new features from additional sources, such as artificial nutrition and blood test results, among others, since their incorporation could yield better performance as well as clinical interpretability. From a machine learning perspective, we are investigating the use of alternative NN architectures as well as distance and similarity methods tailored for MTS and multimodal data. Another relevant line of work is to generalize the proposed model, which currently focuses on predicting the first AMR, to provide a score of the risk of acquiring AMR infections on a day-to-day basis. Such a model would assist in real-time decision-making, enabling clinicians to dynamically adapt the treatment provided to the patients as well as to make on-the-fly (isolation) decisions to prevent the transmission of the AMR in the ICU. Last but not least, while our multimodal architecture was designed to predict and interpret AMR using EHR, MTS and multimodal data are pervasive in contemporary applications. As a result, we are interested in generalizing and adapting our results to other relevant applications, including finance, marketing, and transportation, to name a few.

\section{Availability of Data and Materials}
\label{sec:declaration}
The data used for this research comprise confidential patient health information, which is protected and may not be publicly released. Researchers interested in having access to the data must get approval from the Committee of Ethics of the UHF. The ML and data processing architectures developed in this paper have been programmed in Python, with the associated code being publicly available at \url{https://github.com/smaaguero/MIDDM} (cf. footnote 1).

\section*{Acknowledgements}
This work is supported by the Spanish NSF grants PID2019-106623RB-C41 \hspace{0.2mm} (Big\-Theory),  \hspace{0.2mm} PID2019-\-105032GB-\-I00 \hspace{0.2mm} (SP\-Graph), and PID2019-107768RA-I00 (AAVis-BMR); as well as the Community of Madrid in the framework of the Multiannual Agreement with Rey Juan Carlos University action line ``Young Researchers R\&D Projects'' Refs. F661 (Mapping-UCI) and F861 (AUTO-BA-GRAPH).
S.M. Agüero was awarded with a \hspace{0.15mm} ``URJC \hspace{0.1mm} Predoctoral \hspace{0.1mm} Contracts \hspace{0.1mm} for \hspace{0.1mm} Trainees'' \hspace{0.1mm} grant \hspace{0.2mm} (PREDOC21-036).

\section*{Ethical declarations}
This work was approved by the Research Ethics Committee of the University Hospital of Fuenlabrada (internal reference $16/32$) under the framework of a Spanish Research Project.





% \end{thebibliography}

\printcredits

\bibliographystyle{model1-num}
% \bibliographystyle{cas-model2-names}
% \bibliographystyle{elsarticle-num}

\bibliography{samplebib}

\bio{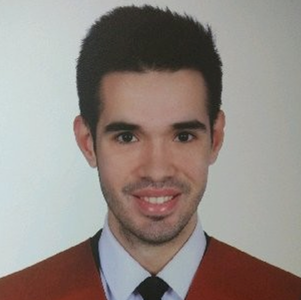}
Sergio Martínez-Agüero (MSc. in Telecommunication Engineering, Rey Juan Carlos University, Spain, 2020) is a Research Assistant at Rey Juan Carlos University currently working on his PhD entitled ``Deep Learning and Network Analytics for extracting knowledge from infectious diseases in the ICU". He has made several contributions to national and international congresses and published two papers in JCR journals. He is currently part of two competitive projects funded by the Spanish Government related to healthcare data-driven ML models. He is interested in data science, ML, data visualization, and network analytics.
\endbio

\bio{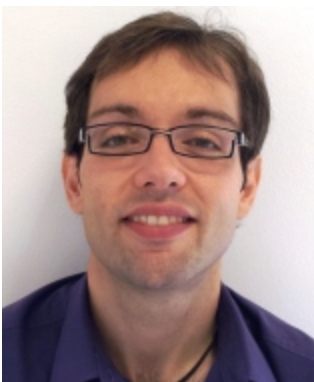}
Antonio G. Marques (PhD. in ECE, Carlos III University of Madrid, Spain, 2007) is a Full Professor at Rey Juan Carlos University, Spain, and  held different visiting positions with the Universities of Minnesota and Pennsylvania, USA.
His current research focuses on signal processing, ML and optimization over graphs and networks. He has served as an Associate Editor and Technical/General Chair for different journals and conferences. His work has been awarded in several venues and he was the recipient of the 2020 EURASIP Early Career Award. Prof. Marques is a Member of the IEEE, EURASIP and the ELLIS society.
\endbio

\bio{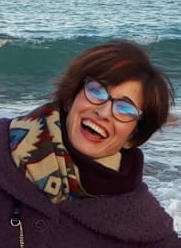}
Inmaculada Mora-Jiménez (PhD. in Telecommunication Engineering, Carlos III University of Madrid, Spain, 2004) is a Full Professor at Rey Juan Carlos University, Spain. She has conducted her research mainly in data analytic and biomedical engineering. She is a co-author of more than 40 JCR-indexed papers and 50 contributions to international conferences. She has participated in 18 competitive research projects (principal investigator of 5) and collaborated in more than 20 projects with private funding entities. Her main research interests include data science and ML with application to image processing, bioengineering, and wireless communications.
\endbio

\bio{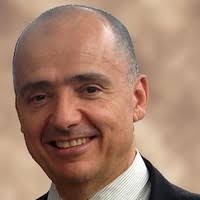}
Joaquín Álvarez-Rodríguez (PhD in Medicine, Complutense University of Madrid, Spain, 1996) has been, since 2003, the head of the Intensive Care Medicine Department at the Hospital Universitario de Fuenlabrada. His lines of work have been the quality and safety of patients, medical information systems and infections in the ICU. He has actively participated in the national coordination of Zero Projects, which aim to reduce the main infections acquired in ICU and the emergence of AMR bacteria in the ICU. His main research area is the collection of data recorded in the electronic medical record.
\endbio

\bio{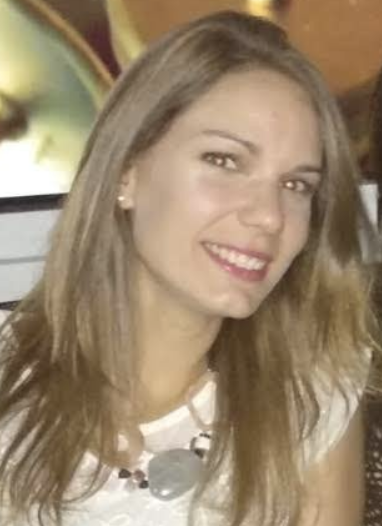}
Cristina Soguero-Ruiz (PhD. in ML with Applications in Healthcare, Rey Juan Carlos University and University Carlos III of Madrid, Spain, 2015) is an Assistant Professor and the Coordinator of the Biomedical Engineering Degree at Rey Juan Carlos University. She won the Orange Foundation Best PhD. Thesis Award by the Spanish Official College of Telecommunication Engineering. She has published more than 30 JRC-indexed papers and 50 international conference communications. She has participated in several research projects related to healthcare data-driven ML systems (being the principal investigator in 5).  Her current research interests include ML and data science.
\endbio

\end{document}